\documentclass[10pt,twocolumn,letterpaper]{article}

\usepackage[pagenumbers]{cvpr} 

\usepackage[T1]{fontenc}

\usepackage{cuted} 

\usepackage{currfile} 

\usepackage{caption} 

\usepackage{multirow}
\usepackage{bm}
\usepackage{colortbl}
\definecolor{famcov}{HTML}{E8F0FA}
\definecolor{famatt}{HTML}{FDEEE3}
\definecolor{famrnd}{HTML}{F1F1F1}
\definecolor{hdrpope}{HTML}{E6F4EA}
\definecolor{hdrgqa}{HTML}{EFE7F6}
\definecolor{hdrgrey}{HTML}{EEF0F2}
\newcommand{\cS}{\mathcal{S}}

\definecolor{cvprblue}{rgb}{0.21,0.49,0.74}
\usepackage[pagebackref,breaklinks,colorlinks,allcolors=cvprblue]{hyperref}

\def\paperID{*****}
\def\confName{CVPR}
\def\confYear{2027}

\title{Confidence under Visual Token Pruning:\\Removed Evidence and Risk-Controlled Token Budgets for MLLMs}

\author{Kaizhen Tan$^{1}$ \quad Yang Feng$^{2}$ \quad Heqing Du$^{2}$ \quad Hanzhe Hong$^{1}$ \quad Siru Tao$^{1}$ \quad Xin Xu$^{1}$\\[0.4em]
$^{1}$Carnegie Mellon University, Pittsburgh, PA, USA \qquad $^{2}$Columbia University, New York, NY, USA}

\begin{document}
\maketitle
\begin{strip}
\centering
\includegraphics[width=\linewidth]{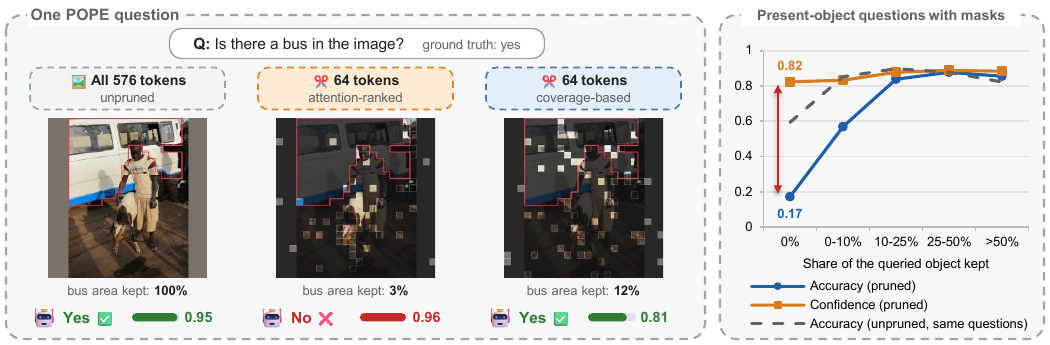}
\captionsetup{hypcap=false}
\captionof{figure}{Left: a POPE question answered by LLaVA-1.5-7B from all 576 visual tokens and from 64 tokens kept by two selectors (red outline: queried object; dimmed cells: pruned tokens; bars: confidence). The attention-ranked selector (CLS attention) keeps 3\% of the bus area, and the model denies the bus with confidence 0.96; the coverage-based selector keeps 12\% and answers correctly. Right: on the 4{,}053 POPE questions about present objects that have a COCO mask (seven selectors, 64 or 128 tokens), accuracy drops as less of the object survives pruning, while confidence changes little.}
\captionsetup{hypcap=true}
\label{fig:teaser}
\end{strip}

\begin{abstract}
Visual token pruning speeds up multimodal large language models (MLLMs) by keeping a small subset of the visual tokens, and pruning methods are compared by the accuracy they retain.
We study what pruning does to the confidence of these models, across common selectors, several MLLMs, and different output formats.
Pruning errors concentrate on questions whose evidence the selector removed, and confidence does not register the removal.
When the queried object loses all of its tokens, accuracy on these questions drops from 59\% to 17\%, while confidence stays at the unpruned level.
Returning a few object tokens to the kept set recovers most of the lost accuracy.
Selectors that keep the most attended tokens remove such evidence most often and produce confident errors, which temperature scaling cannot re-rank.
Selectors that avoid keeping redundant tokens stay close to the calibration of the unpruned model.
We then use the confidence of the pruned model to set a per-question token budget.
The model answers with few tokens first and again with all tokens when its confidence is low.
Conformal risk control sets the threshold to bound the expected deviation from the unpruned model.
With coverage-based selection, this cascade needs about a third of the prefill tokens of the unpruned model, while with FastV it needs more than the unpruned model.
The savings come mainly from how well the confidence ranks the answers that differ from the unpruned ones.
\end{abstract}

\section{Introduction}
\label{sec:intro}

Multimodal large language models (MLLMs) represent an image as a sequence of visual tokens that is usually much longer than the question.
LLaVA-1.5~\citep{liu2023llava} turns every image into 576 tokens, while a short question with its prompt takes about 60 tokens.
Models with native-resolution encoders produce several hundred tokens or more per image~\citep{bai2025qwen25vl}.
Visual token pruning shortens this sequence by keeping a small subset of the visual tokens.
Many training-free selectors now decide which tokens to keep~\citep{chen2024fastv,visionzip2025,deng2025scope,alvar2025divprune,zhang2025cdpruner}, and they are compared by the accuracy they retain at a fixed token budget.

Deployed systems also use the confidence that comes with an answer.
They act on confident answers and abstain~\citep{geifman2017selective} or pass the question to a larger model~\citep{gupta2024cascades} when confidence is low.
These decisions assume calibration, that is, agreement between the confidence of an answer and how often answers with that confidence are correct.
Pruning methods are evaluated by accuracy and cost, and it is unclear whether a pruned model still signals when its answer is likely to be wrong.

The way pruning changes the input suggests a specific failure.
The language model was trained on complete token sets, and after pruning its input contains no indication that tokens are missing.
If the removed tokens carried the evidence for the answer, the model answers from the remaining tokens, and its confidence reflects only what those tokens show.
\Cref{fig:teaser} gives an example from POPE~\citep{li2023pope}, which asks whether an object is present in an image.
A selector that keeps the 64 most attended tokens retains 3\% of the area of a bus that fills much of the image.
LLaVA-1.5 then answers that there is no bus, with confidence 0.96.
A selector that spreads 64 tokens over the image retains 12\% of the bus, and the model answers correctly.

We test this explanation on nine selectors and on five models from two families, LLaVA-1.5 (7B, 13B) and Qwen2.5-VL (3B, 7B, 32B).
The evaluation covers four output formats: yes/no answers (POPE), multiple-choice answers (ScienceQA), open-ended answers scored by their sequence probability (GQA), and object mentions in long image descriptions.
For every question we record which tokens the selector keeps, and on POPE we measure how much of the queried object survives pruning.
With seven selectors at 64 or 128 tokens, the queried object loses all of its tokens in 4{,}557 answers of LLaVA-1.5-7B.
The pruned model answers 17\% of these questions correctly, against 59\% for the unpruned model on the same questions.
Its mean confidence on these questions (0.82) stays at the level of the unpruned model (0.80).
Attention-ranked selectors, which keep the tokens that receive the most attention, remove the queried object more often and barely lower the confidence of the resulting errors.
For two of them, returning up to eight object tokens to the kept set recovers most of the accuracy that the removal cost on these questions.
Returning the same number of other tokens does not.
Coverage- and diversity-based selectors, which avoid keeping redundant tokens, stay close to the calibration of the unpruned model.
On POPE, coverage selection with 128 tokens matches the unpruned accuracy within one point and lowers the expected calibration error (ECE) from 0.040 to 0.017.
On LLaVA-1.5-13B and Qwen2.5-VL with 22\% of the tokens, FastV has the highest ECE on POPE and GQA.
The coverage- and diversity-based selectors raise the ECE of the unpruned model by at most 0.024.
In long descriptions, the removed evidence appears as objects that the pruned model does not mention.
Temperature scaling fitted to each pruned model brings every ECE to 0.07 or less.
As a monotone map, it cannot change how confidence ranks right and wrong answers, and attention-ranked pruning degrades this ranking.
For FastV with 64 tokens, the area under the ROC curve (AUROC) of confidence for correctness falls from 0.801 to 0.674.

The confidence of a pruned model can also decide how many visual tokens a question needs.
Each question is first answered with a small budget and moves to a larger budget only when the confidence falls below a threshold.
We choose the threshold with conformal risk control~\citep{angelopoulos2024crc}, so that the expected share of answers that differ from the unpruned model stays below a target $\varepsilon$.
The bound holds for every selector, and the computation needed to meet it depends on the selector.
We measure this computation by the number of input tokens the language model processes before answering (prefill tokens).
On POPE with $\varepsilon{=}2\%$ and a first budget of 64 tokens, coverage selection needs 36\% of the prefill tokens of the unpruned model and answers 2.3 times faster in batches of 32.
With an uninformative score it would need 97\%, so the saving comes from how well its confidence ranks the answers that differ from the unpruned model.
FastV needs 105\%, more than running the unpruned model directly.

Our contributions are:
\begin{itemize}
\item A per-question study of how visual token pruning changes the confidence of MLLMs, which links pruning errors to the evidence that the selector removes and tests this link by editing the kept tokens.
\item The finding that the selector decides whether this confidence stays reliable.
Attention-ranked selection produces confident errors, and coverage- and diversity-based selection stays close to the calibration of the unpruned model.
\item A confidence-gated token budget with a distribution-free bound on the deviation from the unpruned model, whose cost depends on how well the confidence ranks the disagreements of the pruned model.
\end{itemize}

\section{Related Work}
\label{sec:related}

\paragraph{Visual token pruning.}
Training-free selectors reduce the number of visual tokens of an MLLM and differ mainly in the signal that ranks tokens~\citep{tokencompress2026survey}.
Attention-ranked methods keep the tokens that receive the most attention from the class token of the vision encoder~\citep{visionzip2025} or from text tokens inside the language model~\citep{chen2024fastv,zhang2025sparsevlm,xing2025vico}.
Coverage- and diversity-based methods avoid keeping redundant tokens.
They use facility-location coverage~\citep{deng2025scope,zhu2026coverpruner}, large pairwise distances between kept tokens~\citep{alvar2025divprune,adaptiveprune2026}, a determinantal point process conditioned on the question~\citep{zhang2025cdpruner}, or propagation on a token graph~\citep{jiang2025gprune}.
Adaptive methods choose the budget of each input from the image and the question before the model answers~\citep{atpllava2025}.
All of these methods are evaluated by accuracy and cost.
\citet{wang2025worsethanrandom} report that, beyond about the twentieth layer of the language model, existing training-free selectors perform no better than random selection.
We ask a different question about the same selectors: whether the confidence of the pruned model still indicates when its answer is wrong.

\paragraph{Confidence and uncertainty in MLLMs.}
MLLMs are often overconfident~\citep{chen2025unveiling}.
Existing work improves the confidence of language and multimodal models by training~\citep{conftuner2025} or by a post-hoc map such as temperature scaling~\citep{guo2017calibration}.
Other methods estimate the uncertainty of a model that receives all of its visual tokens by perturbing the inputs~\citep{zhang2024vluncertainty,festa2025} or by sampling several answers~\citep{kuhn2023semantic,umpire2026}.
For convolutional networks, weight pruning can make calibration better or worse depending on the structure and ratio of pruning~\citep{misra2024hidden,mitra2024investigating,brc2025}.
Token pruning leaves the weights unchanged and removes part of the input instead, and its effect on confidence has received little study.

\paragraph{Selective prediction and adaptive computation.}
Selective prediction abstains when confidence is low~\citep{geifman2017selective}, and model cascades send low-confidence inputs to a larger model~\citep{gupta2024cascades}.
CALM~\citep{schuster2022calm} exits early from a language model when a confidence measure is high, and bounds the change of its output with Learn-then-Test~\citep{angelopoulos2025ltt}.
Conformal risk control~\citep{angelopoulos2024crc} bounds the expected value of any monotone loss with a single calibrated threshold.
We apply this principle to the visual token budget of an MLLM.
Unlike adaptive pruning, our budget is decided after a first answer, from its confidence, and the reliability of that confidence depends on which tokens the selector keeps.

\section{Study Setup}
\label{sec:setup}

\paragraph{Pruned inference.}
An MLLM such as LLaVA-1.5~\citep{liu2023llava} encodes an image into $N$ visual tokens and maps them into the embedding space of a language model.
The language model receives these tokens before the tokens of the question.
Visual token pruning keeps a subset $\cS$ of $K<N$ visual tokens.
We remove the other tokens from the input sequence and from the key--value cache, so the language model receives no placeholder for them and no signal that tokens are missing.\footnote{Zeroing the hidden states of pruned tokens differs from removing them, because the zeroed positions still produce keys and receive attention. For FastV at $K{=}128$, zeroing lowers POPE accuracy from 77.8\% to 50.1\% (supplementary material).}
Selectors differ in where they choose $\cS$.
Input-level selectors rank tokens with signals from the vision encoder before the language model runs.
In-LLM selectors let the first layers of the language model process all tokens and remove tokens for the remaining layers.

\paragraph{Selectors.}
\Cref{tab:selectors} lists the nine selectors we study, grouped by the signal that decides which tokens survive.
Attention-ranked selectors keep the tokens that receive the most attention.
CLS attention and VisionZip~\citep{visionzip2025} use the attention of the class token of the vision encoder.
FastV~\citep{chen2024fastv} and PDrop~\citep{xing2025vico} use the attention of the last text token inside the language model.
Coverage- and diversity-based selectors avoid keeping redundant tokens.
SCOPE~\citep{deng2025scope} adds tokens greedily, each time choosing the token with the largest gain in the facility-location coverage
\begin{equation}
C(\cS)=\frac{1}{N}\sum_{u=1}^{N}\max_{s\in\cS}\cos(\mathbf{f}_u,\mathbf{f}_s),
\label{eq:coverage}
\end{equation}
where $\mathbf{f}_u$ is the vision-encoder feature of token $u$.
Each gain is multiplied by the CLS attention of the candidate token raised to a power $\alpha$.
We report the default $\alpha{=}1$ and the variant $\alpha{=}0$, which maximizes coverage alone and which we call coverage selection.
DivPrune~\citep{alvar2025divprune} maximizes the minimum pairwise distance within the kept set.
CDPruner~\citep{zhang2025cdpruner} selects a diverse set with a determinantal point process whose kernel is weighted by the relevance of each token to the question.
Random selection keeps a uniform sample.
For LLaVA-1.5 the budgets are $K\in\{64,128,192\}$ of $N{=}576$ tokens, that is, keep ratios of 11\%, 22\%, and 33\%.
PDrop uses its published three-stage schedules, whose average number of visual tokens per layer equals $K$.

\begin{table}[t]
\centering
\small
\setlength{\tabcolsep}{3.2pt}
\begin{tabular}{@{}lllc@{}}
\toprule
Family & Selector & Ranking signal & Stage \\
\midrule
\multirow{4}{*}{Attention} & CLS attention & class-token attention & input \\
 & VisionZip & class-token attn., merging & input \\
 & FastV & last-token attn., layer 2 & LLM \\
 & PDrop & last-token attn., 3 stages & LLM \\
\midrule
\multirow{4}{*}{\shortstack[l]{Coverage /\\diversity}} & Coverage ($\alpha{=}0$) & facility location, \cref{eq:coverage} & input \\
 & SCOPE ($\alpha{=}1$) & coverage gain $\times$ attn. & input \\
 & DivPrune & max--min distance & input \\
 & CDPruner & question-weighted DPP & input \\
\midrule
 & Random & uniform & input \\
\bottomrule
\end{tabular}
\caption{The nine token selectors, grouped by the signal that ranks tokens. ``Input'' selectors act before the language model; ``LLM'' selectors remove tokens after the first layers of the language model.}
\label{tab:selectors}
\end{table}

\paragraph{Models.}
LLaVA-1.5-7B and 13B~\citep{liu2023llava} use a CLIP ViT-L/14 encoder at 336 pixels~\citep{radford2021clip}, which yields $N{=}576$ visual tokens.
Qwen2.5-VL-3B, 7B, and 32B~\citep{bai2025qwen25vl} encode images at native resolution and merge each $2\times2$ group of patches into one token, so a POPE image gives 360 tokens on average.
On Qwen2.5-VL we run five selectors: coverage selection, DivPrune, attention in the last vision block, FastV, and random selection.
The vision encoder of Qwen2.5-VL has no class token, so attention in its last block takes the place of CLS attention.
The five selectors use the same keep ratios as on LLaVA-1.5, and every kept token keeps its original rotary position.
With all tokens kept, each pruned pipeline reproduces the first answer token of the unpruned model on every checked sample.

\paragraph{Tasks and confidence.}
All models decode greedily with the same prompts in every condition.
POPE~\citep{li2023pope} asks 9{,}000 yes/no questions about the presence of an object in 500 COCO images.
ScienceQA-IMG~\citep{lu2022scienceqa} has 2{,}017 multiple-choice science questions with images.
For these two benchmarks the confidence of an answer is the probability of the predicted option at the first answer token, normalized over the options.
GQA~\citep{hudson2019gqa} asks 12{,}578 open-ended questions (testdev-balanced).
Here the confidence is the probability of the whole generated answer, including the end-of-sequence token, and an answer is correct if it matches the reference after normalization.
To test long-form output, we also ask each model to describe the 500 POPE images in detail.
We extract the objects that a description mentions with the CHAIR protocol~\citep{rohrbach2018chair}.
A mention is correct if the object appears in the COCO annotations or reference captions of the image.
Its confidence is the probability of the first token of the mentioned word.

\paragraph{Metrics.}
We measure calibration with the ECE over 15 equal-width confidence bins~\citep{guo2017calibration}, the gap between mean confidence and accuracy in each bin averaged with bin sizes as weights.
The AUROC measures how well confidence separates correct from wrong answers, independent of its scale.
We also report the confident-error rate, the share of answers with confidence of at least $0.9$ that are wrong.
Because POPE and GQA ask many questions about each image, confidence intervals come from a paired bootstrap that resamples images.
Two one-sided tests with a margin of one point decide whether two accuracies are equivalent.

\paragraph{What a selector removes.}
We use two measures of the evidence that pruning removes.
The first is the coverage $C(\cS)$ of the kept set in \cref{eq:coverage}.
The second is specific to the question.
For each POPE question about an object that is present, we map the COCO instance masks of that object onto the token grid.
Such masks exist for 4{,}053 of the 4{,}500 questions about present objects.
Object retention is the share of the object's area that falls on kept tokens.
Both measures need only the kept indices, which we record for every question.
VisionZip merges part of the tokens and PDrop keeps a different set in each stage, so we compute both measures for the other seven selectors.

\section{How Pruning Changes Confidence}
\label{sec:analysis}

We first compare the selectors on LLaVA-1.5-7B (\cref{sec:selector}) and relate the errors of pruned models to the evidence that the selectors remove (\cref{sec:evidence}).
We then extend the comparison to other models and output formats (\cref{sec:general}) and test what post-hoc recalibration can repair (\cref{sec:recal}).

\begin{table*}[t]
\centering
\small
\setlength{\tabcolsep}{3.8pt}
\begin{tabular}{@{}llcccccccccccc@{}}
\toprule
\cellcolor{hdrgrey} & \cellcolor{hdrgrey} & \multicolumn{6}{c}{\cellcolor{hdrpope}POPE (first-token confidence)} & \multicolumn{6}{c}{\cellcolor{hdrgqa}GQA (sequence confidence)} \\
\cellcolor{hdrgrey} & \cellcolor{hdrgrey} & \multicolumn{3}{c}{\cellcolor{hdrpope}$K{=}64$} & \multicolumn{3}{c}{\cellcolor{hdrpope}$K{=}128$} & \multicolumn{3}{c}{\cellcolor{hdrgqa}$K{=}64$} & \multicolumn{3}{c}{\cellcolor{hdrgqa}$K{=}128$} \\
\cellcolor{hdrgrey}Family & \cellcolor{hdrgrey}Selector & \cellcolor{hdrpope}Acc & \cellcolor{hdrpope}ECE$\downarrow$ & \cellcolor{hdrpope}AUROC & \cellcolor{hdrpope}Acc & \cellcolor{hdrpope}ECE$\downarrow$ & \cellcolor{hdrpope}AUROC & \cellcolor{hdrgqa}Acc & \cellcolor{hdrgqa}ECE$\downarrow$ & \cellcolor{hdrgqa}AUROC & \cellcolor{hdrgqa}Acc & \cellcolor{hdrgqa}ECE$\downarrow$ & \cellcolor{hdrgqa}AUROC \\
\midrule
\multicolumn{2}{@{}l}{Unpruned ($N{=}576$)} & 87.0 & .040 & .801 & 87.0 & .040 & .801 & 62.0 & .053 & .801 & 62.0 & .053 & .801 \\
\midrule
\cellcolor{famatt} & \cellcolor{famatt}CLS attention & 80.5 & .090 & .764 & 84.5 & .050 & .805 & 55.1 & .088 & .765 & 57.9 & .072 & .779 \\
\cellcolor{famatt} & \cellcolor{famatt}VisionZip & 80.7 & .075 & .776 & 84.8 & .046 & .807 & 55.1 & .087 & .767 & 57.6 & .073 & .778 \\
\cellcolor{famatt} & \cellcolor{famatt}FastV & 71.1 & .214 & .674 & 77.8 & .146 & .715 & 51.6 & .115 & .763 & 55.7 & .098 & .773 \\
\cellcolor{famatt}\multirow{-4}{*}{\shortstack[l]{Attention-ranked}} & \cellcolor{famatt}PDrop & 69.1 & .173 & .735 & 81.5 & .085 & .798 & 48.8 & .105 & .754 & 55.0 & .093 & .773 \\
\midrule
\cellcolor{famcov} & \cellcolor{famcov}Coverage ($\alpha{=}0$) & 85.1 & .025 & .810 & 87.0 & .017 & .814 & 58.1 & .064 & .777 & 59.1 & .056 & .786 \\
\cellcolor{famcov} & \cellcolor{famcov}SCOPE ($\alpha{=}1$) & 85.4 & .032 & .815 & 87.0 & .023 & .818 & 58.5 & .064 & .772 & 59.5 & .058 & .782 \\
\cellcolor{famcov} & \cellcolor{famcov}DivPrune & 86.6 & .009 & .812 & 87.5 & \textbf{.013} & .812 & 57.6 & \textbf{.060} & .782 & 59.2 & \textbf{.051} & \textbf{.787} \\
\cellcolor{famcov}\multirow{-4}{*}{\shortstack[l]{Coverage /\\diversity}} & \cellcolor{famcov}CDPruner & \textbf{88.0} & \textbf{.007} & \textbf{.819} & \textbf{87.9} & .015 & \textbf{.819} & \textbf{58.5} & .071 & \textbf{.785} & \textbf{59.8} & .062 & .787 \\
\midrule
\cellcolor{famrnd} & \cellcolor{famrnd}Random & 79.5 & .071 & .773 & 83.6 & .046 & .790 & 55.3 & .082 & .763 & 58.2 & .064 & .774 \\
\bottomrule
\end{tabular}
\caption{Selectors on LLaVA-1.5-7B with 64 and 128 of the 576 visual tokens: accuracy (\%), ECE, and AUROC. POPE uses the first-token confidence and GQA the probability of the whole answer. Bold: best pruned selector per column.}
\label{tab:main}
\end{table*}

\subsection{Calibration Depends on the Selector}
\label{sec:selector}

\Cref{tab:main} reports accuracy, ECE, and AUROC for every selector at 64 and 128 tokens.
On POPE, the attention-ranked selectors lose accuracy and become overconfident.
With 64 tokens their ECE ranges from 0.075 (VisionZip) to 0.214 (FastV), against 0.040 for the unpruned model, and their AUROC falls from 0.801 to between 0.674 and 0.776.
FastV is the extreme case: 24\% of its answers with confidence of at least 0.9 are wrong, compared with 5.7\% for the unpruned model.
The coverage- and diversity-based selectors keep ECE below the unpruned level at both budgets (0.007 to 0.032) and keep AUROC above it (0.810 to 0.819).
Some of them also keep accuracy.
With 128 tokens, coverage selection and SCOPE are equivalent to the unpruned model within one point.
For coverage selection the difference is $+0.1$ (95\% interval $[-0.7,0.8]$), and the ECE falls to 0.017.
CDPruner with 64 tokens is one point more accurate than the unpruned model (interval $[0.3,1.8]$) and has an ECE of 0.007.
Random selection falls between the two families.

Accuracy alone does not determine these differences.
FastV with 192 tokens and random selection with 64 tokens answer a similar share of POPE correctly (81.4\% and 79.5\%).
FastV has an ECE of 0.109 and an AUROC of 0.729, and random selection has 0.071 and 0.773.

On GQA, where the confidence is the probability of the whole generated answer, the selectors keep the same order.
With 64 tokens the attention-ranked selectors reach ECE between 0.087 and 0.115 and the coverage- and diversity-based selectors between 0.060 and 0.071.
With 128 tokens the two ranges are 0.072 to 0.098 and 0.051 to 0.062.
GQA also shows a limit of pruning: no pruned model is better calibrated than the unpruned one at 64 tokens (0.053), and only DivPrune matches it at 128 tokens (0.051).
On this open-ended task every selector answers between 48.8\% and 59.8\% of the questions correctly, against 62.0\% without pruning, and the lower confidence of the pruned models does not make up for it.

\subsection{Pruning Errors Come from Removed Evidence}
\label{sec:evidence}

\begin{figure*}[t]
\centering
\includegraphics[width=\textwidth]{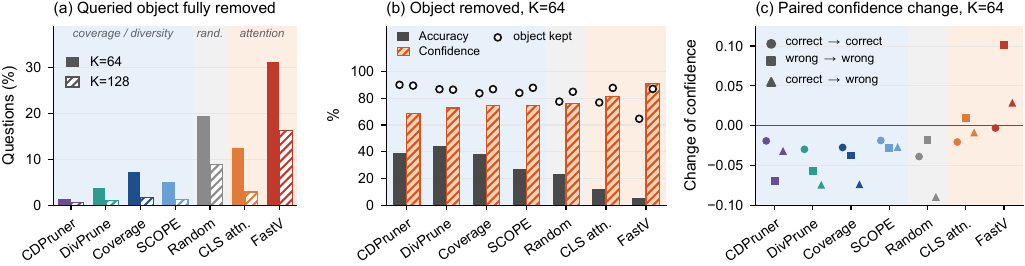}
\caption{Removed evidence on POPE (LLaVA-1.5-7B, questions about present objects). (a) Share of questions whose object loses all of its tokens. (b) Accuracy and mean confidence on these questions; ticks: questions whose object keeps some tokens. (c) Mean change of confidence on the same question from the unpruned to the pruned model.}
\label{fig:evidence}
\end{figure*}

To find where the errors of pruned models come from, we measure for every POPE question with a present object how much of the object's area survives pruning (\cref{sec:setup}).
\Cref{fig:evidence}a shows how often the object loses all of its tokens.
With 64 tokens, FastV removes the queried object completely for 31\% of these questions and CLS attention for 12\%.
The coverage- and diversity-based selectors do so for 1\% to 7\%.
The coverage $C(\cS)$ of the whole image does not explain these rates.
CLS attention keeps the lowest coverage (0.53) and coverage selection the highest (0.88), but the kept sets of FastV and CDPruner with 64 tokens have the same coverage (0.70).
FastV removes the queried object for 31\% of the questions and CDPruner for 1.3\%.

\Cref{fig:evidence}b shows what happens when the evidence is gone.
Accuracy on these questions drops to 5\% for FastV and 12\% for CLS attention, and to between 23\% and 44\% for the other selectors.
Confidence changes much less.
It stays between 0.69 and 0.91, and for FastV it is higher on questions whose object was removed (0.91) than on questions whose object kept some tokens (0.87).
Across all seven selectors at 64 and 128 tokens, wrong answers whose object was removed have a mean confidence of 0.85, and the other wrong answers 0.77.

On the same questions (\cref{fig:evidence}c), the coverage- and diversity-based selectors lower confidence more on answers that stay wrong ($-0.03$ to $-0.07$) than on answers that stay correct ($-0.02$ to $-0.03$).
This asymmetry is consistent with their lower ECE.
FastV raises confidence by 0.10 on answers that stay wrong and by 0.03 on answers that it turns wrong.

Changing the kept set directly shows that the removal causes these errors (supplementary material).
For the questions whose object CLS attention or FastV removed with 64 tokens, we replace up to eight kept tokens with the tokens that cover the largest parts of the object.
Accuracy rises from 12.5\% to 61.7\% for CLS attention and from 5.0\% to 64.6\% for FastV, for which the object tokens are put back at the second layer.
The unpruned model answers 57.1\% and 68.1\% of these questions correctly.
Replacing the same kept tokens with tokens outside the object leaves accuracy at 13.3\% and 5.2\%.
Conversely, removing the object tokens from the kept sets of coverage selection lowers accuracy from 83.3\% to 33.1\%, while the mean confidence falls only from 0.87 to 0.78.
These measurements match the explanation in \cref{sec:intro}.
Accuracy depends on whether the evidence for the question survives, and confidence does not track this.
A pruned model therefore stays calibrated when its selector rarely removes the evidence that a question needs.

\subsection{Other Models and Output Formats}
\label{sec:general}

\begin{table}[t]
\centering
\small
\setlength{\tabcolsep}{3.6pt}
\begin{tabular}{@{}lcccccc@{}}
\toprule
Model & \cellcolor{hdrgrey}Unpr. & \cellcolor{famcov}Cov. & \cellcolor{famcov}DivP. & \cellcolor{famatt}Attn. & \cellcolor{famatt}FastV & \cellcolor{famrnd}Rand. \\
\midrule
\multicolumn{7}{@{}l}{\cellcolor{hdrpope}\emph{POPE, ECE$\downarrow$}} \\
LLaVA-1.5-13B & .032 & .017 & \textbf{.017} & .054 & .107 & .052 \\
Qwen2.5-VL-3B & .067 & .074 & \textbf{.063} & .080 & .094 & .078 \\
Qwen2.5-VL-7B & .082 & \textbf{.084} & .086 & .097 & .126 & .107 \\
Qwen2.5-VL-32B & .085 & .109 & \textbf{.102} & .103 & .188 & .125 \\
\midrule
\multicolumn{7}{@{}l}{\cellcolor{hdrgqa}\emph{GQA, ECE$\downarrow$}} \\
LLaVA-1.5-13B & .060 & .067 & \textbf{.064} & .081 & .093 & .080 \\
Qwen2.5-VL-3B & .134 & .124 & .115 & .145 & .181 & \textbf{.085} \\
Qwen2.5-VL-7B & .200 & .199 & \textbf{.193} & .218 & .254 & .194 \\
Qwen2.5-VL-32B & .161 & \textbf{.171} & .174 & .181 & .217 & .188 \\
\bottomrule
\end{tabular}
\caption{Other models at a keep ratio of 22\%. For LLaVA-1.5-13B, Attn.\ is CLS attention; for Qwen2.5-VL, it is the attention received in the last vision block. GQA uses the probability of the whole answer. Bold: best pruned selector in each row.}
\label{tab:models}
\end{table}

\Cref{tab:models} repeats the comparison on LLaVA-1.5-13B and on Qwen2.5-VL at a keep ratio of 22\%.
On every model, FastV has the highest ECE on both POPE and GQA.
Coverage selection and DivPrune raise the ECE of the unpruned model by at most 0.024, and on LLaVA-1.5 they lower it on POPE.
The removed-evidence pattern also holds on Qwen2.5-VL.
With 11\% of the tokens, FastV removes the queried object for 18\% of the POPE questions about present objects on Qwen2.5-VL-7B and for 52\% on Qwen2.5-VL-32B.
The two models answer 5\% and 18\% of these questions correctly.
Their mean confidence on these questions (0.93 and 0.84) equals their mean confidence on questions whose object keeps some tokens.

On ScienceQA-IMG, pruning changes accuracy and calibration little.
LLaVA-1.5-7B stays within 1.4 points of its unpruned accuracy (63.8\%) with every selector and budget.
Without the image it still answers 58.9\% of the questions correctly, so the image adds about five points to its accuracy.
On Qwen2.5-VL, pruning costs up to 8.6 points of accuracy, and no selector raises the ECE by more than 0.02 (supplementary material).

In long descriptions, removed evidence appears as omission.
With 64 tokens, the attention-ranked selectors mention 15\% to 34\% fewer objects than the unpruned LLaVA-1.5-7B.
The share of ground-truth objects they mention falls from 77\% to between 51\% and 65\%.
Their share of hallucinated mentions stays at 24\% (unpruned: 25\%).
Coverage- and diversity-based selectors mention 72\% to 76\% of the ground-truth objects and hallucinate slightly more (27\% to 28\%; supplementary material).
In a description, the omission of a removed object corresponds to answering ``no'' to a question about it.

\subsection{What Recalibration Repairs}
\label{sec:recal}

\begin{table}[t]
\centering
\small
\setlength{\tabcolsep}{3.4pt}
\begin{tabular}{@{}lccccc@{}}
\toprule
 & \multicolumn{3}{c}{\cellcolor{hdrgrey}ECE$\downarrow$} & \multicolumn{2}{c}{\cellcolor{hdrgrey}AUROC$\uparrow$} \\
Selector & raw & $T_{\text{unpr.}}$ & $T_{\text{own}}$ & raw & +answer \\
\midrule
\cellcolor{hdrgrey}Unpruned & .040 & .014 & .014 & .801 & .848 \\
\midrule
\cellcolor{famcov}Coverage & .025 & .023 & .017 & .810 & .843 \\
\cellcolor{famcov}DivPrune & .009 & .038 & .011 & .812 & .835 \\
\cellcolor{famcov}CDPruner & .007 & .039 & .008 & .819 & .821 \\
\cellcolor{famatt}CLS attention & .090 & .049 & .038 & .764 & .829 \\
\cellcolor{famatt}VisionZip & .075 & .035 & .034 & .776 & .834 \\
\cellcolor{famatt}PDrop & .173 & .124 & .066 & .735 & .813 \\
\cellcolor{famatt}FastV & .214 & .178 & .041 & .674 & .832 \\
\bottomrule
\end{tabular}
\caption{Post-hoc recalibration on POPE (LLaVA-1.5-7B, 64 tokens), fitted on half of the images and evaluated on the other half. $T_{\text{unpr.}}$: temperature fitted to the unpruned model; $T_{\text{own}}$: temperature fitted to each pruned model; both leave AUROC unchanged. +answer: logistic recalibration that also receives the predicted answer.}
\label{tab:recal}
\end{table}

\Cref{tab:recal} tests three ways to recalibrate a pruned model on POPE with 64 tokens, each fitted on half of the images and evaluated on the other half.
The temperature of the unpruned model, which a system that adds pruning later would reuse, does not transfer.
It leaves FastV at an ECE of 0.178 and PDrop at 0.124 and raises the ECE of DivPrune and CDPruner from below 0.01 to about 0.04.
A temperature fitted to each pruned model brings every ECE to 0.07 or less.
Both are monotone maps and leave AUROC unchanged, so the pruned models still rank their wrong answers as before: 0.674 for FastV against 0.801 for the unpruned model.
A logistic recalibration that also receives the predicted answer raises the AUROC of every selector to at least 0.81.
It can do so because the errors of pruned models on POPE are mostly ``no'' answers to questions about present objects.
This repair relies on the yes/no format of POPE and has no direct counterpart for open-ended answers.

Sampling-based uncertainty does not improve the ranking either.
On 3{,}000 GQA questions, the agreement of ten sampled answers separates correct from wrong answers less well than the probability of the greedy answer in all six configurations we tested (supplementary material).

\section{Risk-Controlled Token Budgets}
\label{sec:method}

\Cref{sec:analysis} shows that the confidence of a pruned model separates correct from wrong answers when the selector keeps the evidence that questions need.
It separates them much less when the selector keeps only the most attended tokens.
We now use this confidence to decide how many visual tokens each question receives, with a decision rule that bounds the loss relative to the unpruned model.

\begin{figure}[t]
\centering
\includegraphics[width=\columnwidth]{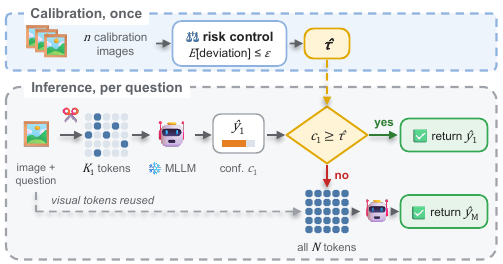}
\caption{Confidence-gated token budget with two stages. Conformal risk control chooses $\hat\tau$ once on calibration images. Each question is answered with $K_1$ visual tokens and, if its confidence is below $\hat\tau$, again with all tokens, reusing the visual tokens of the first pass.}
\label{fig:method}
\end{figure}

\paragraph{Cascade.}
Fix a selector and budgets $K_1<K_2<\dots<K_M$, where $K_M=N$ is the unpruned model.
For a question $x$, let $\hat y_m(x)$ and $c_m(x)$ be the answer and its confidence (as defined in \cref{sec:setup}) when the selector keeps $K_m$ tokens.
Given a threshold $\tau$, the cascade runs the stages in order and returns the answer of the first stage whose confidence reaches $\tau$:
\begin{equation}
\hat y_\tau(x)=\hat y_{m^\ast}(x),\ \ m^\ast=\min\{m{<}M\!: c_m(x){\ge}\tau\}.
\label{eq:cascade}
\end{equation}
If no pruned stage reaches $\tau$, the cascade returns the answer of the unpruned model (\cref{fig:method}).
Raising $\tau$ sends more questions to later stages and costs more tokens.
The cost of a question is the sum of the prefill tokens of all stages it passes through.
A question that ends at the unpruned model therefore costs more than running the unpruned model directly.

\paragraph{Loss.}
We measure the effect of pruning by the deviation from the unpruned model,
$\ell(x;\tau)=\mathbf{1}[\hat y_\tau(x)\neq\hat y_M(x)]$.
This loss needs no labels.
A deviation rate of $\varepsilon$ also bounds the accuracy loss relative to the unpruned model by $\varepsilon$.
The cascade can turn a correct answer into a wrong one only where the two answers differ.

\paragraph{Choosing the threshold.}
Benchmarks such as POPE and GQA ask many questions about one image, and these questions are not independent.
We therefore treat images as the exchangeable units.
For an image $j$ with question set $Q_j$, let $L_j(\tau)=\frac{1}{|Q_j|}\sum_{x\in Q_j}\ell(x;\tau)$.
The threshold takes values in a fixed grid $\Lambda$, whose largest value $\tau_{\max}$ exceeds every confidence, so that at $\tau_{\max}$ every question reaches the unpruned model.
With more than two stages, $L_j$ can increase with $\tau$, because a later stage may deviate where an earlier stage agreed.
Following \citet{angelopoulos2024crc}, we therefore work with the upper envelope $\tilde L_j(\tau)=\max_{\tau'\in\Lambda,\,\tau'\ge\tau}L_j(\tau')$, which does not increase with $\tau$ and is zero at $\tau_{\max}$.
Given $n$ calibration images and a target $\varepsilon$, conformal risk control selects
\begin{equation}
\hat\tau=\min\Big\{\tau\in\Lambda:\ \frac{1}{n+1}\sum_{j=1}^{n}\tilde L_j(\tau)+\frac{1}{n+1}\le\varepsilon\Big\}.
\label{eq:crc}
\end{equation}
If the calibration images and a new test image are exchangeable, Corollary~1 of \citet{angelopoulos2024crc} for losses bounded by one gives
\begin{equation}
\mathbb{E}\big[L_{n+1}(\hat\tau)\big]\le\varepsilon ,
\end{equation}
where the expectation is over the calibration images and the test image.
The bound concerns the deviation of an image averaged over its questions.

\paragraph{Role of the confidence.}
The bound holds for any selector and any confidence score, and the cost at which it is met depends on both.
It depends on how often the first stage disagrees with the unpruned model and on how well the confidence ranks these disagreements below the agreements.
If the pruned model is confident mainly when it agrees with the unpruned model, $\hat\tau$ is low and most questions stop at the first stage.
If it deviates with high confidence, only a high $\hat\tau$ keeps these deviations below $\varepsilon$, and most questions reach the unpruned model.

\section{Experiments with Risk-Controlled Budgets}
\label{sec:experiments}

\paragraph{Setup.}
We evaluate the cascade of \cref{sec:method} with two stages.
In the first stage the selector keeps 11\% of the visual tokens ($K{=}64$ on LLaVA-1.5).
Questions whose confidence falls below the threshold are answered again with all tokens.
We fix this design before looking at calibration data, so that choosing among designs does not weaken the bound.
Other designs are compared in the supplementary material.
For each configuration we split the images at random into a calibration half and a test half.
We select $\hat\tau$ by \cref{eq:crc} on the calibration half, from a grid uniform in the logit of $\tau$.
Deviation and cost are measured on the test half and averaged over 200 splits.
Cost is the number of prefill tokens relative to the unpruned model.
An escalated question pays for both stages, and the first stage of FastV includes its two full-length layers.
To separate the two factors of \cref{sec:method}, we also run the cascade with two other scores.
An uninformative random score gives no ranking, and a perfect gate escalates exactly the questions on which the first stage disagrees with the unpruned model.
As a budget chosen before answering, an input-adaptive baseline uses the smallest $K\in\{64,128,192\}$ whose kept set reaches a coverage $C(\cS)\ge c$.
It uses all tokens if no budget does, and $c$ is selected by the same risk-control procedure.

\paragraph{Validity.}
Averaged over the test splits, the deviation from the unpruned model, taken per image as in \cref{sec:method}, stays below $\varepsilon$ on POPE and GQA in every configuration we evaluate.
At $\varepsilon{=}2\%$ it is at most 1.7\%, and the deviation pooled over all questions and the loss of accuracy also stay within $\varepsilon$.

\paragraph{Cost.}
On POPE with LLaVA-1.5-7B and $\varepsilon{=}2\%$, the coverage- and diversity-based selectors need between 32\% and 36\% of the prefill tokens of the unpruned model.
The attention-ranked selectors need between 65\% and 105\%, and PDrop and FastV need about as much as the unpruned model or more.
For every $\varepsilon$ from 0.5\% to 10\%, the coverage- and diversity-based selectors are the cheapest on POPE and FastV the most expensive (supplementary material).

With 64 tokens, the coverage- and diversity-based selectors disagree with the unpruned model on 5\% to 7\% of the POPE questions.
A random score cannot single out these few disagreements and must escalate most questions.
Meeting the bound with it would still take 89\% to 97\% of the prefill tokens, and a perfect gate would take 25\% to 27\%.
Their confidence separates disagreements from agreements with an AUROC of 0.90 to 0.93 and realizes 85\% to 89\% of this possible saving.
FastV disagrees on 21\% of the questions, and its confidence (AUROC 0.67) realizes 18\%.
VisionZip and CLS attention disagree about equally often (11\% and 12\%).
The confidence of VisionZip ranks the disagreements better (AUROC 0.81 against 0.78), and its cascade costs 0.65 instead of 0.73.
Re-scoring the answers of FastV does not close the gap.
We fit a logistic gate on the confidence and the predicted answer to agreement with the unpruned model, on a separate quarter of the images.
On these splits, which leave a quarter of the images for calibration, the raw confidence costs 1.08 for FastV and 0.39 for coverage selection.
The learned gate lowers the cost of FastV to 0.91 (supplementary material).

On GQA the savings are smaller because pruning changes more answers: the coverage- and diversity-based selectors disagree on 18\% to 20\% of the questions.
At $\varepsilon{=}2\%$ they need between 72\% and 80\% of the prefill tokens, and the attention-ranked selectors between 89\% and 108\%.
With the four coverage- and diversity-based selectors, the input-adaptive budget needs 70\% to 77\% of the prefill tokens on POPE, about twice as much as the cascade.
It relies on the coverage of the whole image, which does not show whether the evidence of a particular question survives (\cref{sec:evidence}).

\begin{table}[t]
\centering
\small
\setlength{\tabcolsep}{2.1pt}
\begin{tabular}{@{}llcccccc@{}}
\toprule
Model & Bench. & \cellcolor{famcov}Cov. & \cellcolor{famcov}DivP. & \cellcolor{famcov}CDP. & \cellcolor{famatt}Attn. & \cellcolor{famatt}FastV & \cellcolor{famrnd}Rand. \\
\midrule
LLaVA-1.5-7B & POPE & 0.36 & 0.34 & \textbf{0.32} & 0.73 & 1.05 & 0.65 \\
 & GQA & 0.80 & 0.80 & \textbf{0.72} & 0.89 & 1.08 & 0.93 \\
LLaVA-1.5-13B & POPE & 0.38 & 0.36 & \textbf{0.27} & 0.72 & 0.97 & 0.65 \\
 & GQA & 0.87 & 0.88 & \textbf{0.78} & 0.94 & 1.03 & 0.97 \\
Qwen2.5-VL-3B & POPE & \textbf{0.45} & 0.50 & -- & 0.62 & 0.85 & 0.68 \\
 & GQA & \textbf{0.86} & 0.97 & -- & 1.01 & 1.11 & 1.04 \\
Qwen2.5-VL-7B & POPE & \textbf{0.41} & 0.48 & -- & 0.62 & 0.95 & 0.63 \\
 & GQA & \textbf{0.77} & 0.89 & -- & 0.99 & 1.13 & 0.99 \\
Qwen2.5-VL-32B & POPE & \textbf{0.52} & 0.55 & -- & 0.75 & 0.98 & 0.70 \\
 & GQA & \textbf{0.87} & 0.93 & -- & 0.99 & 1.13 & 0.97 \\
\bottomrule
\end{tabular}
\caption{Prefill cost (unpruned model $=1$) of the two-stage cascade at $\varepsilon{=}2\%$, with 11\% of the visual tokens in the first stage. Attn.: CLS attention on LLaVA-1.5, last-block attention on Qwen2.5-VL. Bold: cheapest selector.}
\label{tab:budget}
\end{table}

\paragraph{Other models.}
\Cref{tab:budget} reports the cost at $\varepsilon{=}2\%$ for all models.
On POPE and GQA, a coverage- or diversity-based selector is the cheapest on every model, and FastV is the most expensive; on GQA it exceeds the unpruned cost on all five models.

\paragraph{Wall-clock time.}
We time the cascade with coverage selection on all 9{,}000 POPE questions in batches of 32 on an A100 GPU.
The time includes the vision encoder, the selection, both prefill stages, and the readout.
The unpruned model takes 477\,s and the cascade 207\,s, a speedup of 2.3.
The cascade escalates 17.5\% of the questions, and their second pass reuses the visual tokens of the first.
Because the vision encoder (40\,s) and the selection (6\,s) do not shrink, the speedup is smaller than the factor of 2.8 given by the prefill tokens.
At batch size one the saving is small: on an H200 GPU, a POPE question takes 53\,ms without pruning and 45 to 72\,ms with the pruned models (supplementary material).

\section{Conclusion}
\label{sec:conclusion}

Pruning errors of MLLMs concentrate on questions whose evidence the selector removed, and confidence does not register the removal.
Attention-ranked selectors remove this evidence most often and produce confident errors, while coverage- and diversity-based selectors stay close to the calibration of the unpruned model.
The same difference decides the cost of a confidence-gated token budget with a bounded deviation.
Its savings come from how well the confidence ranks the answers that differ from the unpruned model.
We therefore suggest reporting calibration next to accuracy and cost when pruning methods are compared.
Our evidence measurements rely on COCO masks, so for open-ended answers we observe the pattern only through calibration and omitted objects.
The bound assumes exchangeable images.

{
    \small
    \bibliographystyle{ieeenat_fullname}
    \bibliography{references}
}

\clearpage
\maketitlesupplementary
\appendix
\renewcommand{\topfraction}{0.95}
\renewcommand{\bottomfraction}{0.9}
\renewcommand{\textfraction}{0.05}
\renewcommand{\floatpagefraction}{0.8}
\renewcommand{\dbltopfraction}{0.95}
\renewcommand{\dblfloatpagefraction}{0.8}
\setcounter{topnumber}{4}
\setcounter{bottomnumber}{2}
\setcounter{totalnumber}{6}
\setcounter{dbltopnumber}{4}
\makeatletter
\setlength{\@fptop}{0pt}
\setlength{\@fpsep}{\floatsep}
\setlength{\@fpbot}{0pt plus 1fil}
\setlength{\@dblfptop}{0pt}
\setlength{\@dblfpsep}{\dblfloatsep}
\setlength{\@dblfpbot}{0pt plus 1fil}
\makeatother
\section{Implementation Details}
\label{supp:impl}

\paragraph{Token removal.}
Every input-level selector receives the projected visual tokens of LLaVA-1.5 and returns the indices it keeps.
We slice the input embeddings at these indices, keep the order of the kept tokens, and remove the other positions from the attention mask.
FastV runs the first two decoder layers on the full sequence, ranks the visual tokens by the attention of the last prompt token at the second layer (averaged over heads), and removes the other visual tokens from the hidden states and from the key--value cache of every later layer; kept tokens keep their original rotary positions.
PDrop uses the implementation distributed with SCOPE. After the 2nd, 6th, and 16th decoder layers it keeps $(66,30,17)$, $(303,110,36)$, or $(300,200,110)$ visual tokens for the budgets $K{=}64$, $128$, and $192$, so that the average number of visual tokens per layer is close to $K$.
VisionZip keeps $K{-}c$ dominant tokens and $c$ merged contextual tokens with $c{=}10$, $20$, and $30$.

\paragraph{Zeroing pruned tokens.}
A shortcut for FastV sets the hidden states of the pruned tokens to zero after the ranking layer instead of removing them.
The zeroed positions still produce keys with attention logit $0$ rather than $-\infty$, so after the softmax several hundred of them take a large share of every attention distribution.
In our audit this shortcut lowered POPE accuracy of LLaVA-1.5-7B at $K{=}128$ to $50.1\%$, chance level, while the removal implementation reached $77.8\%$.
Our FastV implementation reproduces the unpruned model when no token is removed (identical first-token argmax on all checked samples; largest logit difference $0.016$ in half precision).

\paragraph{Qwen2.5-VL.}
We splice the merged vision tokens into the text embeddings, select tokens at the language-model input, and pass the original three-dimensional rotary positions of the kept tokens.
Decoding runs through the decoder layers with our own loop so that the input-level and FastV paths share one numerical path; we use the same loop with all tokens as the unpruned reference.
With all tokens kept, both paths reproduce the first-token argmax of \texttt{generate} on all 30 checked questions.
Images are resized by the model processor with at most $1280\times28\times28$ pixels.

\section{Additional Results}
\label{supp:results}

\paragraph{Full grids.}
\Cref{tab:supp_grid_llava-v15-7b-a,tab:supp_grid_llava-v15-7b-b,tab:supp_grid_llava-v15-13b,tab:supp_grid_qwen25-vl-3b,tab:supp_grid_qwen25-vl-7b,tab:supp_grid_qwen25-vl-32b} list every selector and budget for every model and benchmark, including the keep ratio of 33\% that the main text omits.
For GQA we report both the first-token confidence and the probability of the whole answer.

\begin{figure*}[t]
\centering
\includegraphics[width=\textwidth]{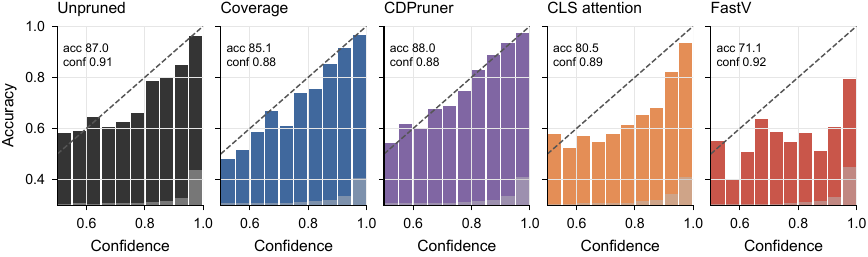}
\caption{Reliability diagrams on POPE (LLaVA-1.5-7B, 64 tokens). Bars: accuracy per confidence bin; gray: share of answers per bin; dashed: perfect calibration.}
\label{fig:supp_reliability}
\end{figure*}

\paragraph{Reliability diagrams.}
\Cref{fig:supp_reliability} shows the accuracy in each confidence bin on POPE with 64 tokens.
Coverage selection and CDPruner lie close to the diagonal.
CLS attention and FastV keep most answers at high confidence while their accuracy in those bins falls, which produces their large ECE.

\begin{figure}[t]
\centering
\includegraphics[width=\columnwidth]{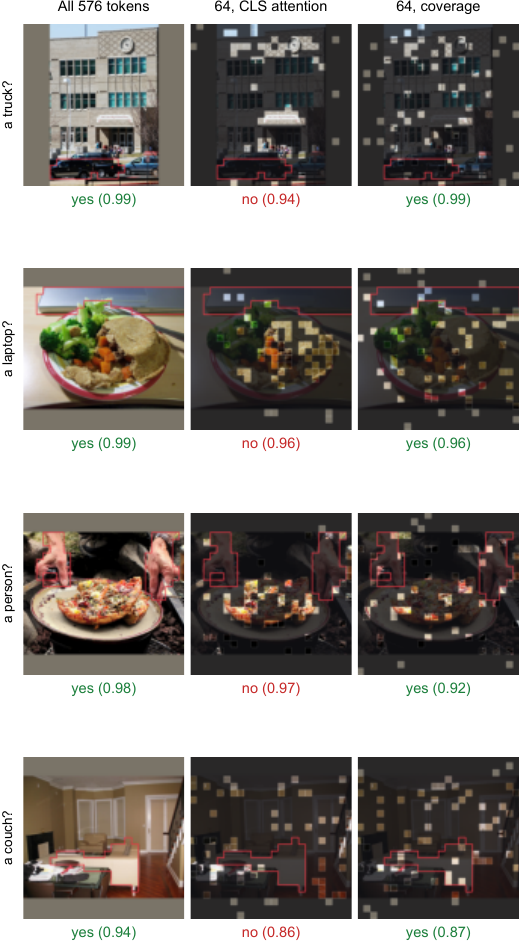}
\caption{POPE questions answered by LLaVA-1.5-7B with all tokens and with 64 tokens kept by CLS attention or coverage selection. Red outline: queried object; dimmed cells: pruned tokens. Below each panel: answer and confidence (green: correct, red: wrong).}
\label{fig:supp_examples}
\end{figure}

\paragraph{Examples.}
\Cref{fig:supp_examples} shows four more POPE questions in which CLS attention with 64 tokens removes the queried object and the pruned model denies it with confidence between 0.86 and 0.97, while coverage selection keeps part of the object and the answer stays correct.

\paragraph{Object-token intervention.}
\Cref{tab:supp_intervention} gives the intervention of Sec.~4.2 of the main paper on the questions about present objects with an available mask.
Let $O$ be the set of grid cells that overlap the queried object.
For CLS attention and FastV with 64 tokens we use the 504 and 1{,}260 questions whose object lost all of its tokens.
We replace $r=\min(|O|,8)$ randomly chosen kept tokens with the $r$ cells of $O$ that hold the largest object area; for FastV the replacement happens at the second layer, where FastV removes tokens.
The control replaces the same kept tokens with $r$ random cells that are neither kept nor in $O$.
For coverage selection we use the 3{,}567 questions whose object keeps between 1 and 32 cells.
We replace all kept cells of $O$ with random cells outside $O$, and the control replaces the same number of kept cells outside $O$.
The budget stays at 64 tokens in every condition, and the random choices are seeded by the question.
Rerunning the recorded kept sets reproduces the recorded answers on 99.4\% (CLS attention) and 100\% (FastV, coverage selection) of the questions.
Restoring at most eight cells brings back 91\% (CLS attention) and 87\% (FastV) of the object area on average.
Relative to the control on the same questions, accuracy changes by $+48.4$ points (95\% interval $[41.5,55.9]$, image-clustered bootstrap) for CLS attention, $+59.4$ $[54.8,64.1]$ for FastV, and $-50.2$ $[-53.0,-47.3]$ for coverage selection.
The mean confidence changes by $-0.04$, $-0.07$, and $-0.09$.

\paragraph{Long descriptions and sampling.}
\Cref{tab:supp_caption} lists the object-level results of the detailed descriptions and \cref{tab:supp_sampling} the comparison with sampling-based uncertainty on GQA.

\begin{figure}[t]
\centering
\includegraphics[width=\columnwidth]{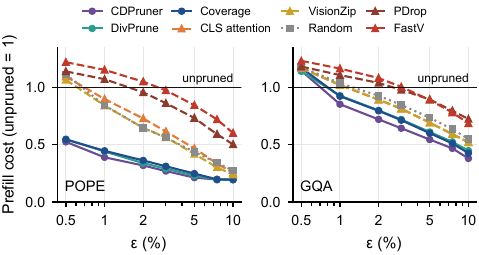}
\caption{Prefill cost at which the cascade keeps the expected deviation from the unpruned LLaVA-1.5-7B below $\varepsilon$ (first stage: 64 tokens). Solid: coverage- and diversity-based selectors; dashed: attention-ranked selectors; dotted: random.}
\label{fig:supp_budget}
\end{figure}

\paragraph{Cost across targets.}
\Cref{fig:supp_budget} shows the cost of the two-stage cascade on LLaVA-1.5-7B for targets $\varepsilon$ from 0.5\% to 10\%.

\paragraph{Where the savings come from.}
\Cref{tab:supp_gate} decomposes the cost of the two-stage cascade at $\varepsilon{=}2\%$.
An uninformative score (independent uniform values) has to escalate about a share $1-\varepsilon/D$ of the questions, where $D$ is the disagreement rate of the first stage.
A perfect gate escalates exactly the disagreements and costs the first stage plus $D$.
Across the five models, the confidence of the coverage- and diversity-based selectors realizes 71\% to 95\% of the saving between the two on POPE and 31\% to 54\% on GQA; that of FastV realizes 18\% to 43\% and 18\% to 25\%.
\Cref{tab:supp_learned} lists a label-free learned gate on POPE.
A logistic model predicts agreement with the unpruned answer from the logit of the confidence, an indicator of a ``no'' answer, and their product.
It is fitted on a quarter of the images, the threshold is selected by conformal risk control on another quarter, and cost and deviation are measured on the remaining half; the raw confidence is evaluated on the same splits.
The learned gate lowers the cost of the attention-ranked selectors, random selection, and CDPruner, and leaves the other coverage- and diversity-based selectors unchanged; FastV and PDrop still need about 91\% of the prefill tokens.

\paragraph{Cascade designs.}
\Cref{tab:supp_cascade} compares cascade designs on LLaVA-1.5-7B at $\varepsilon{=}2\%$.
Passing through all three budgets before the unpruned model costs more than the two-stage cascade that starts at 11\%, for every selector and benchmark, because each stage adds its prefill tokens and later stages add deviations of their own.
Starting at 22\% instead of 11\% of the tokens lowers the cost of the attention-ranked selectors and of random selection on POPE, for example from 0.73 to 0.55 for CLS attention.
It raises the cost of the coverage- and diversity-based selectors, whose first stage at 11\% already answers more than 80\% of the POPE questions.

\paragraph{Validity on ScienceQA-IMG.}
In ScienceQA-IMG every question comes with its own image, so every question is a calibration unit, and the correction term $1/(n{+}1)$ in Eq.~(3) of the main paper is about $0.001$.
The average deviation is then close to $\varepsilon$.
Over 200 splits, the average test deviation of the two-stage cascades that start at 11\% lies within one standard error (about $0.0004$) of $\varepsilon{=}2\%$, in some configurations slightly above it and in others slightly below.
Across all designs and targets, the average exceeds $\varepsilon$ by at most $0.0006$.
On POPE and GQA, with 250 and 199 calibration images, the correction term is $0.004$ to $0.005$, and the average deviation stays below $\varepsilon$ in every configuration.

\paragraph{Latency and throughput.}
\Cref{tab:supp_timing} lists the median time to answer one POPE question at batch size one on an H200 GPU.
The unpruned model takes 52.7\,ms, and FastV, VisionZip, and PDrop save 4 to 7\,ms.
Our input-level selectors run the vision encoder a second time to obtain the features and the class-token attention used for selection.
This pass is included in their latency, and together with the greedy selection it outweighs the saving (random selection: 52.5\,ms; coverage selection with 128 tokens: 72.2\,ms).
\Cref{tab:supp_e2e} times complete pipelines on all 9{,}000 POPE questions in batches of 32 on an A100 GPU, from the preprocessed image to the answer.
The cascade uses coverage selection with 64 tokens and the threshold $\tau{=}0.735$, the median of the thresholds selected on the calibration halves at $\varepsilon{=}2\%$.
It escalates 17.5\% of the questions, whose second pass reuses the visual tokens of the first, and its answers differ from the unpruned ones on 1.4\% of the questions.
The batched pipelines reproduce the answers of the batch-1 runs on 99.99\% (unpruned) and 98.8\% (64 tokens) of the questions.
\Cref{tab:supp_throughput} measures the prefill of the language model alone on an A100 GPU.
With 32 or more questions per batch, 64 visual tokens give 5.0 times the throughput of 576 visual tokens, close to the 5.1-fold reduction of input tokens.
At batch size one the gain is 2.2 times.

\begin{table}[t]
\centering
\small
\setlength{\tabcolsep}{3pt}
\begin{tabular}{@{}llccccc@{}}
\toprule
Selector & Kept set & $n$ & Retention & Acc & Conf. & ``no'' \\
\midrule
CLS attention & recorded & 504 & 0.0 & 12.5 & .816 & 87.5 \\
 & restored &  & 90.7 & 61.7 & .770 & 38.3 \\
 & control &  & 0.0 & 13.3 & .812 & 86.7 \\
\midrule
FastV & recorded & 1260 & 0.0 & 5.0 & .910 & 95.0 \\
 & restored &  & 87.0 & 64.6 & .825 & 35.4 \\
 & control &  & 0.0 & 5.2 & .898 & 94.8 \\
\midrule
Coverage & recorded & 3567 & 20.0 & 83.3 & .866 & 16.7 \\
 & removed &  & 0.0 & 33.1 & .776 & 66.9 \\
 & control &  & 20.0 & 83.2 & .866 & 16.8 \\
\bottomrule
\end{tabular}
\caption{Object-token intervention on POPE (LLaVA-1.5-7B, 64 tokens, questions about present objects). Retention: share of the object area on kept tokens (\%). Acc and ``no'': share of correct and of ``no'' answers (\%).}
\label{tab:supp_intervention}
\end{table}

\begin{table}[t]
\centering
\small
\setlength{\tabcolsep}{3pt}
\begin{tabular}{@{}lcccccc@{}}
\toprule
 & \multicolumn{3}{c}{ECE} & \multicolumn{3}{c}{AUROC} \\
\cmidrule(lr){2-4}\cmidrule(l){5-7}
Condition & first & seq. & agree$_{10}$ & first & seq. & agree$_{10}$ \\
\midrule
CLS attention, 128 & .083 & .072 & .090 & .768 & .773 & .736 \\
CLS attention, 64 & .100 & .088 & .104 & .753 & .760 & .733 \\
Coverage, 128 & .066 & .056 & .093 & .762 & .771 & .730 \\
Coverage, 64 & .076 & .064 & .086 & .764 & .772 & .733 \\
DivPrune, 128 & .062 & .053 & .075 & .771 & .778 & .748 \\
Unpruned & .057 & .048 & .085 & .773 & .782 & .748 \\
\bottomrule
\end{tabular}
\caption{Confidence scores on the first 3{,}000 GQA questions (LLaVA-1.5-7B): first-token probability, probability of the greedy answer, and agreement of ten answers sampled at temperature one with the greedy answer.}
\label{tab:supp_sampling}
\end{table}

\begin{table}[t]
\centering
\small
\setlength{\tabcolsep}{4pt}
\begin{tabular}{@{}lcccc@{}}
\toprule
 & \multicolumn{2}{c}{Latency (ms)} & \multicolumn{2}{c}{Selection (ms)} \\
\cmidrule(lr){2-3}\cmidrule(l){4-5}
Selector & 64 & 128 & 64 & 128 \\
\midrule
Unpruned & \multicolumn{2}{c}{52.7} & \multicolumn{2}{c}{--} \\
\midrule
CLS attention & 53.6 & 52.4 & 0.2 & 0.2 \\
VisionZip & 46.0 & 45.3 & -- & -- \\
FastV & 45.4 & 45.3 & -- & -- \\
PDrop & 48.6 & 48.3 & -- & -- \\
Coverage ($\alpha{=}0$) & 62.2 & 72.2 & 9.8 & 19.0 \\
DivPrune & 56.4 & 61.0 & 4.2 & 8.5 \\
CDPruner & 65.1 & 70.1 & 11.8 & 17.6 \\
Random & 52.5 & 52.5 & 0.1 & 0.1 \\
\bottomrule
\end{tabular}
\caption{Median time to answer one POPE question with LLaVA-1.5-7B at batch size one on an H200 GPU, including the vision encoder and decoding, and mean time of the token selection itself. Selection time is listed for the input-level selectors that run a separate selection step.}
\label{tab:supp_timing}
\end{table}

\begin{table}[t]
\centering
\small
\setlength{\tabcolsep}{5pt}
\begin{tabular}{@{}lcccc@{}}
\toprule
 & \multicolumn{4}{c}{Visual tokens} \\
\cmidrule(l){2-5}
Batch & 64 & 128 & 192 & 576 \\
\midrule
1 & 44.9 & 41.6 & 37.6 & 20.6 \\
8 & 97.5 & 70.5 & 53.3 & 21.9 \\
32 & 110.5 & 73.0 & 55.5 & 22.1 \\
64 & 112.3 & 74.8 & 56.1 & 22.2 \\
\bottomrule
\end{tabular}
\caption{Prefill throughput (questions per second) of the language model of LLaVA-1.5-7B on an A100 GPU for 61 text tokens and a given number of visual tokens. The vision encoder is excluded because pruning does not change its cost.}
\label{tab:supp_throughput}
\end{table}

\begin{table}[t]
\centering
\small
\setlength{\tabcolsep}{3pt}
\begin{tabular}{@{}lcccccc@{}}
\toprule
Pipeline & Vision & Select & Prefill & Prefill$_{\mathrm{esc}}$ & Total & q/s \\
\midrule
Unpruned & 40 & 0 & 436 & 0 & 477 & 18.9 \\
64 tokens & 40 & 6 & 84 & 0 & 130 & 69.0 \\
Cascade & 40 & 6 & 84 & 76 & 207 & 43.5 \\
\bottomrule
\end{tabular}
\caption{End-to-end time (s) for the 9000 POPE questions with LLaVA-1.5-7B in batches of 32 on an A100 GPU. Select: coverage selection of 64 tokens; Prefill$_{\mathrm{esc}}$: second pass of the escalated questions with all tokens; q/s: questions per second.}
\label{tab:supp_e2e}
\end{table}

\begin{table}[t]
\centering
\small
\setlength{\tabcolsep}{3.5pt}
\begin{tabular}{@{}lcccc@{}}
\toprule
 & \multicolumn{2}{c}{AUROC$_{\mathrm{agr}}$} & \multicolumn{2}{c}{Cost} \\
\cmidrule(lr){2-3}\cmidrule(l){4-5}
Selector & conf. & learned & conf. & learned \\
\midrule
CLS attention & .777 & .818 & 0.78 & 0.68 \\
VisionZip & .807 & .837 & 0.70 & 0.63 \\
FastV & .667 & .802 & 1.08 & 0.91 \\
PDrop & .747 & .801 & 0.99 & 0.91 \\
Coverage ($\alpha{=}0$) & .921 & .921 & 0.39 & 0.39 \\
SCOPE ($\alpha{=}1$) & .904 & .905 & 0.40 & 0.40 \\
DivPrune & .923 & .924 & 0.37 & 0.37 \\
CDPruner & .931 & .969 & 0.34 & 0.28 \\
Random & .817 & .847 & 0.70 & 0.65 \\
\bottomrule
\end{tabular}
\caption{Label-free learned gate on POPE (LLaVA-1.5-7B, 64 tokens, $\varepsilon{=}2\%$). The gate is fitted on a quarter of the images, calibrated on another quarter, and evaluated on the remaining half, averaged over 200 splits; the raw confidence uses the same calibration and test images.}
\label{tab:supp_learned}
\end{table}

\begin{table*}[t]
\centering
\small
\setlength{\tabcolsep}{5pt}
\begin{tabular}{@{}lrrrrrr@{}}
\toprule
Run & Mentions & Recall & CHAIR$_i$ & CHAIR$_s$ & ECE & AUROC \\
\midrule
CDPruner, 128 & 2326 & 76.8 & 28.5 & 66.8 & .194 & .696 \\
CDPruner, 64 & 2304 & 76.1 & 28.5 & 67.2 & .198 & .687 \\
CLS attention, 128 & 2032 & 71.7 & 23.6 & 54.4 & .242 & .706 \\
CLS attention, 64 & 1864 & 65.0 & 24.5 & 54.1 & .241 & .692 \\
Coverage, 128 & 2284 & 76.3 & 27.6 & 65.4 & .209 & .694 \\
Coverage, 64 & 2139 & 72.3 & 26.7 & 62.8 & .202 & .696 \\
DivPrune, 128 & 2334 & 77.3 & 28.3 & 66.8 & .207 & .681 \\
DivPrune, 64 & 2194 & 74.3 & 26.7 & 60.4 & .217 & .682 \\
FastV, 128 & 1693 & 63.3 & 19.1 & 41.9 & .270 & .671 \\
FastV, 64 & 1458 & 51.3 & 23.7 & 45.2 & .254 & .666 \\
Random, 128 & 2075 & 71.1 & 25.8 & 56.4 & .223 & .693 \\
Random, 64 & 1946 & 65.0 & 27.6 & 56.8 & .209 & .671 \\
Unpruned & 2205 & 76.6 & 24.7 & 60.6 & .230 & .705 \\
VisionZip, 128 & 2053 & 72.5 & 23.5 & 54.3 & .245 & .714 \\
VisionZip, 64 & 1833 & 64.4 & 23.9 & 51.0 & .247 & .686 \\
\bottomrule
\end{tabular}
\caption{Object mentions in detailed descriptions of the 500 POPE images (LLaVA-1.5-7B). Recall: share of ground-truth objects mentioned. CHAIR$_i$: share of mentions that are hallucinated; CHAIR$_s$: share of descriptions with at least one hallucination. ECE and AUROC of the first-token probability of each mention.}
\label{tab:supp_caption}
\end{table*}

\begin{table*}[t]
\centering
\footnotesize
\setlength{\tabcolsep}{3.6pt}
\renewcommand{\arraystretch}{0.94}
\begin{tabular}{@{}llcccccccccccc@{}}
\toprule
 & & \multicolumn{6}{c}{\cellcolor{hdrpope}POPE} & \multicolumn{6}{c}{\cellcolor{hdrgqa}GQA} \\
 & &  & & \multicolumn{3}{c}{cost with score} &  &  & & \multicolumn{3}{c}{cost with score} &  \\
\cmidrule(lr){5-7}\cmidrule(lr){11-13}
Model & Selector & $D$ & AUROC$_{\mathrm{agr}}$ & conf. & random & perfect & share & $D$ & AUROC$_{\mathrm{agr}}$ & conf. & random & perfect & share \\
\midrule
LLaVA-1.5-7B & \cellcolor{famatt}CLS attention & 11.8 & .777 & 0.73 & 1.07 & 0.31 & 44.8 & 24.7 & .777 & 0.89 & 1.14 & 0.45 & 35.7 \\
 & \cellcolor{famatt}VisionZip & 11.4 & .807 & 0.65 & 1.06 & 0.31 & 55.1 & 24.6 & .775 & 0.89 & 1.15 & 0.44 & 35.9 \\
 & \cellcolor{famatt}FastV & 21.0 & .667 & 1.05 & 1.18 & 0.46 & 17.7 & 32.0 & .757 & 1.08 & 1.20 & 0.57 & 19.2 \\
 & \cellcolor{famatt}PDrop & 23.3 & .747 & 0.96 & 1.13 & 0.43 & 24.5 & 36.4 & .750 & 1.04 & 1.16 & 0.56 & 20.2 \\
 & \cellcolor{famcov}Coverage & 6.9 & .921 & 0.36 & 0.97 & 0.26 & 86.2 & 18.8 & .795 & 0.80 & 1.11 & 0.39 & 43.7 \\
 & \cellcolor{famcov}SCOPE & 5.8 & .904 & 0.35 & 0.90 & 0.25 & 84.5 & 18.6 & .802 & 0.77 & 1.11 & 0.38 & 47.2 \\
 & \cellcolor{famcov}DivPrune & 6.2 & .923 & 0.34 & 0.94 & 0.26 & 88.0 & 19.6 & .793 & 0.80 & 1.12 & 0.39 & 43.9 \\
 & \cellcolor{famcov}CDPruner & 5.4 & .931 & 0.32 & 0.89 & 0.25 & 89.1 & 17.7 & .822 & 0.72 & 1.11 & 0.38 & 52.7 \\
 & \cellcolor{famrnd}Random & 12.9 & .817 & 0.65 & 1.07 & 0.32 & 56.8 & 25.0 & .773 & 0.93 & 1.13 & 0.45 & 29.8 \\
\midrule
LLaVA-1.5-13B & \cellcolor{famatt}CLS attention & 11.8 & .776 & 0.72 & 1.07 & 0.31 & 45.5 & 25.4 & .762 & 0.94 & 1.14 & 0.45 & 29.9 \\
 & \cellcolor{famatt}VisionZip & 11.5 & .803 & 0.67 & 1.07 & 0.31 & 51.9 & 24.8 & .765 & 0.93 & 1.14 & 0.45 & 30.4 \\
 & \cellcolor{famatt}FastV & 15.9 & .724 & 0.97 & 1.16 & 0.40 & 25.2 & 27.6 & .769 & 1.03 & 1.19 & 0.52 & 25.0 \\
 & \cellcolor{famcov}Coverage & 7.1 & .902 & 0.38 & 0.97 & 0.27 & 83.8 & 20.0 & .773 & 0.87 & 1.12 & 0.40 & 34.7 \\
 & \cellcolor{famcov}DivPrune & 6.0 & .913 & 0.36 & 0.93 & 0.26 & 84.7 & 21.0 & .775 & 0.88 & 1.12 & 0.41 & 34.4 \\
 & \cellcolor{famcov}CDPruner & 4.2 & .941 & 0.27 & 0.81 & 0.24 & 94.7 & 18.2 & .801 & 0.78 & 1.11 & 0.38 & 44.7 \\
 & \cellcolor{famrnd}Random & 13.6 & .826 & 0.65 & 1.08 & 0.33 & 57.5 & 26.0 & .761 & 0.97 & 1.14 & 0.46 & 25.6 \\
\midrule
Qwen2.5-VL-3B & \cellcolor{famatt}CLS attention & 9.7 & .826 & 0.62 & 1.09 & 0.34 & 62.8 & 29.6 & .777 & 1.01 & 1.20 & 0.54 & 28.5 \\
 & \cellcolor{famatt}FastV & 13.3 & .775 & 0.85 & 1.18 & 0.42 & 43.1 & 33.7 & .747 & 1.11 & 1.24 & 0.62 & 21.1 \\
 & \cellcolor{famcov}Coverage & 7.8 & .906 & 0.45 & 1.07 & 0.32 & 82.6 & 21.3 & .811 & 0.86 & 1.17 & 0.46 & 43.4 \\
 & \cellcolor{famcov}DivPrune & 8.4 & .876 & 0.50 & 1.06 & 0.32 & 76.6 & 26.0 & .779 & 0.97 & 1.18 & 0.50 & 31.3 \\
 & \cellcolor{famrnd}Random & 12.0 & .820 & 0.68 & 1.12 & 0.36 & 57.7 & 29.9 & .759 & 1.04 & 1.20 & 0.54 & 24.1 \\
\midrule
Qwen2.5-VL-7B & \cellcolor{famatt}CLS attention & 9.3 & .840 & 0.62 & 1.06 & 0.33 & 61.3 & 27.9 & .783 & 0.99 & 1.19 & 0.52 & 30.4 \\
 & \cellcolor{famatt}FastV & 14.9 & .750 & 0.95 & 1.20 & 0.44 & 32.3 & 36.1 & .746 & 1.13 & 1.26 & 0.66 & 21.4 \\
 & \cellcolor{famcov}Coverage & 6.8 & .910 & 0.41 & 1.02 & 0.31 & 85.5 & 19.3 & .832 & 0.77 & 1.16 & 0.44 & 53.6 \\
 & \cellcolor{famcov}DivPrune & 8.3 & .888 & 0.48 & 1.04 & 0.32 & 78.4 & 24.4 & .809 & 0.89 & 1.18 & 0.49 & 42.7 \\
 & \cellcolor{famrnd}Random & 10.5 & .831 & 0.63 & 1.08 & 0.35 & 61.8 & 27.5 & .786 & 0.99 & 1.19 & 0.52 & 28.8 \\
\midrule
Qwen2.5-VL-32B & \cellcolor{famatt}CLS attention & 12.2 & .792 & 0.75 & 1.12 & 0.36 & 48.9 & 29.9 & .783 & 0.99 & 1.20 & 0.54 & 31.7 \\
 & \cellcolor{famatt}FastV & 24.7 & .755 & 0.98 & 1.20 & 0.51 & 32.6 & 41.5 & .737 & 1.13 & 1.23 & 0.68 & 18.2 \\
 & \cellcolor{famcov}Coverage & 8.3 & .866 & 0.52 & 1.06 & 0.32 & 73.3 & 22.5 & .818 & 0.87 & 1.18 & 0.47 & 44.4 \\
 & \cellcolor{famcov}DivPrune & 9.8 & .860 & 0.55 & 1.09 & 0.34 & 71.2 & 26.8 & .801 & 0.93 & 1.20 & 0.51 & 39.1 \\
 & \cellcolor{famrnd}Random & 12.8 & .826 & 0.70 & 1.12 & 0.37 & 56.8 & 28.1 & .792 & 0.97 & 1.19 & 0.53 & 33.2 \\
\bottomrule
\end{tabular}
\caption{Two-stage cascade at $\varepsilon{=}2\%$ (first stage: 11\% of the visual tokens). $D$: share of questions (\%) on which the first stage disagrees with the unpruned model; AUROC$_{\mathrm{agr}}$: AUROC of the confidence for agreement. Cost with the confidence, with an uninformative random score, and with a gate that escalates exactly the disagreements; share: part of the saving between the random and the perfect score that the confidence reaches (\%).}
\label{tab:supp_gate}
\end{table*}

\begin{table*}[t]
\centering
\small
\setlength{\tabcolsep}{3.2pt}
\begin{tabular}{@{}lcccccccccccc@{}}
\toprule
 & \multicolumn{4}{c}{POPE} & \multicolumn{4}{c}{GQA} & \multicolumn{4}{c}{ScienceQA-IMG} \\
\cmidrule(lr){2-5}\cmidrule(lr){6-9}\cmidrule(lr){10-13}
Selector & 11\% & 22\% & M & IA & 11\% & 22\% & M & IA & 11\% & 22\% & M & IA \\
\midrule
CLS attention & 0.73 & 0.55 & 0.95 & 0.80 & 0.89 & 0.86 & 1.32 & 0.94 & 0.65 & 0.66 & 0.93 & 0.80 \\
VisionZip & 0.65 & 0.52 & 0.83 & -- & 0.89 & 0.87 & 1.33 & -- & 0.66 & 0.63 & 0.93 & -- \\
FastV & 1.05 & 1.04 & 1.57 & -- & 1.08 & 1.07 & 1.69 & -- & 0.73 & 0.62 & 1.04 & -- \\
PDrop & 0.96 & 0.70 & 1.12 & -- & 1.04 & 1.01 & 1.53 & -- & 0.82 & 0.80 & 1.25 & -- \\
Coverage ($\alpha{=}0$) & 0.36 & 0.40 & 0.44 & 0.77 & 0.80 & 0.78 & 1.18 & 0.93 & 0.68 & 0.72 & 1.03 & 0.93 \\
SCOPE ($\alpha{=}1$) & 0.35 & 0.39 & 0.42 & 0.71 & 0.77 & 0.78 & 1.13 & 0.92 & 0.62 & 0.65 & 0.90 & 0.89 \\
DivPrune & 0.34 & 0.41 & 0.40 & 0.75 & 0.80 & 0.79 & 1.17 & 0.93 & 0.68 & 0.71 & 0.99 & 0.91 \\
CDPruner & 0.32 & 0.39 & 0.37 & 0.74 & 0.72 & 0.73 & 1.05 & 0.93 & 0.63 & 0.70 & 0.97 & 0.93 \\
Random & 0.65 & 0.57 & 0.81 & 0.84 & 0.93 & 0.91 & 1.37 & 0.94 & 0.78 & 0.76 & 1.16 & 0.91 \\
\bottomrule
\end{tabular}
\caption{Prefill cost relative to the unpruned LLaVA-1.5-7B at $\varepsilon{=}2\%$. 11\%, 22\%: two-stage cascade whose first stage keeps this share of the visual tokens; M: all three budgets in increasing order before the unpruned model; IA: input-adaptive budget chosen from the coverage of the kept set.}
\label{tab:supp_cascade}
\end{table*}

\begin{table*}[t]
\centering
\footnotesize
\setlength{\tabcolsep}{3.4pt}
\renewcommand{\arraystretch}{0.88}
\begin{tabular}{@{}lcccccccccccc@{}}
\toprule
 & \multicolumn{4}{c}{keep 11\%} & \multicolumn{4}{c}{keep 22\%} & \multicolumn{4}{c}{keep 33\%} \\
\cmidrule(lr){2-5}\cmidrule(lr){6-9}\cmidrule(lr){10-13}
Selector & Acc & ECE & AUROC & CER$_{.9}$ & Acc & ECE & AUROC & CER$_{.9}$ & Acc & ECE & AUROC & CER$_{.9}$ \\
\midrule
\multicolumn{13}{@{}l}{\cellcolor{hdrpope}\emph{POPE, first-token confidence}} \\
\cellcolor{hdrgrey}Unpruned & 87.0 & .040 & .801 & .057 & 87.0 & .040 & .801 & .057 & 87.0 & .040 & .801 & .057 \\
\cellcolor{famatt}CLS attention & 80.5 & .090 & .764 & .097 & 84.5 & .050 & .805 & .063 & 86.4 & .038 & .805 & .055 \\
\cellcolor{famatt}VisionZip & 80.7 & .075 & .776 & .088 & 84.8 & .046 & .807 & .056 & 86.5 & .033 & .812 & .048 \\
\cellcolor{famatt}FastV & 71.1 & .214 & .674 & .237 & 77.8 & .146 & .715 & .158 & 81.4 & .109 & .729 & .127 \\
\cellcolor{famatt}PDrop & 69.1 & .173 & .735 & .166 & 81.5 & .085 & .798 & .085 & 85.6 & .057 & .787 & .068 \\
\cellcolor{famcov}Coverage ($\alpha{=}0$) & 85.1 & .025 & .810 & .046 & 87.0 & .017 & .814 & .041 & 87.6 & .016 & .816 & .042 \\
\cellcolor{famcov}SCOPE ($\alpha{=}1$) & 85.4 & .032 & .815 & .049 & 87.0 & .023 & .818 & .042 & 87.3 & .026 & .819 & .043 \\
\cellcolor{famcov}DivPrune & 86.6 & .009 & .812 & .037 & 87.5 & .013 & .812 & .038 & 87.7 & .017 & .814 & .043 \\
\cellcolor{famcov}CDPruner & 88.0 & .007 & .819 & .034 & 87.9 & .015 & .819 & .036 & 87.8 & .022 & .818 & .041 \\
\cellcolor{famrnd}Random & 79.5 & .071 & .773 & .081 & 83.6 & .046 & .790 & .065 & 85.1 & .040 & .807 & .057 \\
\midrule
\multicolumn{13}{@{}l}{\cellcolor{hdrgqa}\emph{GQA, probability of the whole answer}} \\
\cellcolor{hdrgrey}Unpruned & 62.0 & .053 & .801 & .068 & 62.0 & .053 & .801 & .068 & 62.0 & .053 & .801 & .068 \\
\cellcolor{famatt}CLS attention & 55.1 & .088 & .765 & .113 & 57.9 & .072 & .779 & .096 & 59.4 & .063 & .787 & .077 \\
\cellcolor{famatt}VisionZip & 55.1 & .087 & .767 & .111 & 57.6 & .073 & .778 & .097 & 59.1 & .063 & .786 & .079 \\
\cellcolor{famatt}FastV & 51.6 & .115 & .763 & .172 & 55.7 & .098 & .773 & .126 & 57.9 & .085 & .783 & .107 \\
\cellcolor{famatt}PDrop & 48.8 & .105 & .754 & .151 & 55.0 & .093 & .773 & .110 & 57.9 & .075 & .781 & .092 \\
\cellcolor{famcov}Coverage ($\alpha{=}0$) & 58.1 & .064 & .777 & .086 & 59.1 & .056 & .786 & .072 & 59.7 & .054 & .792 & .075 \\
\cellcolor{famcov}SCOPE ($\alpha{=}1$) & 58.5 & .064 & .772 & .089 & 59.5 & .058 & .782 & .079 & 60.1 & .057 & .791 & .075 \\
\cellcolor{famcov}DivPrune & 57.6 & .060 & .782 & .082 & 59.2 & .051 & .787 & .067 & 60.1 & .050 & .791 & .068 \\
\cellcolor{famcov}CDPruner & 58.5 & .071 & .785 & .086 & 59.8 & .062 & .787 & .081 & 60.4 & .060 & .791 & .076 \\
\cellcolor{famrnd}Random & 55.3 & .082 & .763 & .106 & 58.2 & .064 & .774 & .090 & 59.4 & .056 & .781 & .082 \\
\bottomrule
\end{tabular}
\caption{LLaVA-1.5-7B: every selector and keep ratio (POPE, first-token confidence; GQA, probability of the whole answer). Accuracy in \%, ECE, AUROC, and the share of answers with confidence of at least 0.9 that are wrong (CER$_{.9}$). The unpruned row is repeated in every column group. Cell colours mark the selector families.}
\label{tab:supp_grid_llava-v15-7b-a}
\end{table*}

\begin{table*}[t]
\centering
\footnotesize
\setlength{\tabcolsep}{3.4pt}
\renewcommand{\arraystretch}{0.88}
\begin{tabular}{@{}lcccccccccccc@{}}
\toprule
 & \multicolumn{4}{c}{keep 11\%} & \multicolumn{4}{c}{keep 22\%} & \multicolumn{4}{c}{keep 33\%} \\
\cmidrule(lr){2-5}\cmidrule(lr){6-9}\cmidrule(lr){10-13}
Selector & Acc & ECE & AUROC & CER$_{.9}$ & Acc & ECE & AUROC & CER$_{.9}$ & Acc & ECE & AUROC & CER$_{.9}$ \\
\midrule
\multicolumn{13}{@{}l}{\cellcolor{hdrgqa}\emph{GQA, first-token confidence}} \\
\cellcolor{hdrgrey}Unpruned & 62.0 & .065 & .792 & .079 & 62.0 & .065 & .792 & .079 & 62.0 & .065 & .792 & .079 \\
\cellcolor{famatt}CLS attention & 55.1 & .100 & .758 & .127 & 57.9 & .084 & .770 & .110 & 59.4 & .075 & .778 & .090 \\
\cellcolor{famatt}VisionZip & 55.1 & .099 & .760 & .127 & 57.6 & .085 & .769 & .112 & 59.1 & .075 & .777 & .093 \\
\cellcolor{famatt}FastV & 51.6 & .126 & .756 & .183 & 55.7 & .110 & .765 & .137 & 57.9 & .097 & .775 & .119 \\
\cellcolor{famatt}PDrop & 48.8 & .115 & .747 & .166 & 55.0 & .105 & .766 & .121 & 57.9 & .088 & .773 & .103 \\
\cellcolor{famcov}Coverage ($\alpha{=}0$) & 58.1 & .076 & .769 & .100 & 59.1 & .068 & .778 & .085 & 59.7 & .066 & .784 & .089 \\
\cellcolor{famcov}SCOPE ($\alpha{=}1$) & 58.5 & .076 & .763 & .103 & 59.5 & .071 & .773 & .093 & 60.1 & .069 & .782 & .089 \\
\cellcolor{famcov}DivPrune & 57.6 & .072 & .776 & .092 & 59.2 & .064 & .779 & .082 & 60.1 & .063 & .783 & .083 \\
\cellcolor{famcov}CDPruner & 58.5 & .085 & .776 & .101 & 59.8 & .075 & .777 & .096 & 60.4 & .073 & .781 & .088 \\
\cellcolor{famrnd}Random & 55.3 & .093 & .755 & .119 & 58.2 & .076 & .765 & .102 & 59.4 & .068 & .774 & .096 \\
\midrule
\multicolumn{13}{@{}l}{\cellcolor{hdrgrey}\emph{ScienceQA-IMG, first-token confidence}} \\
\cellcolor{hdrgrey}Unpruned & 63.8 & .180 & .795 & .154 & 63.8 & .180 & .795 & .154 & 63.8 & .180 & .795 & .154 \\
\cellcolor{famatt}CLS attention & 64.4 & .157 & .795 & .129 & 64.1 & .165 & .792 & .117 & 63.2 & .178 & .797 & .137 \\
\cellcolor{famatt}VisionZip & 64.6 & .159 & .798 & .124 & 64.2 & .168 & .793 & .126 & 63.5 & .176 & .798 & .131 \\
\cellcolor{famatt}FastV & 64.6 & .183 & .788 & .155 & 64.6 & .181 & .784 & .149 & 64.1 & .181 & .798 & .151 \\
\cellcolor{famatt}PDrop & 64.3 & .166 & .784 & .146 & 65.2 & .161 & .786 & .137 & 63.9 & .175 & .795 & .154 \\
\cellcolor{famcov}Coverage ($\alpha{=}0$) & 64.8 & .159 & .785 & .136 & 64.4 & .169 & .797 & .124 & 64.1 & .173 & .795 & .135 \\
\cellcolor{famcov}SCOPE ($\alpha{=}1$) & 64.7 & .162 & .791 & .128 & 63.8 & .171 & .800 & .120 & 63.3 & .178 & .803 & .136 \\
\cellcolor{famcov}DivPrune & 63.1 & .178 & .808 & .134 & 64.0 & .171 & .804 & .128 & 63.8 & .174 & .797 & .136 \\
\cellcolor{famcov}CDPruner & 63.7 & .181 & .794 & .140 & 63.7 & .179 & .797 & .133 & 64.0 & .176 & .788 & .142 \\
\cellcolor{famrnd}Random & 63.3 & .168 & .774 & .152 & 63.8 & .173 & .776 & .157 & 63.2 & .178 & .788 & .144 \\
\bottomrule
\end{tabular}
\caption{LLaVA-1.5-7B: every selector and keep ratio (GQA, first-token confidence; ScienceQA-IMG, first-token confidence). Accuracy in \%, ECE, AUROC, and the share of answers with confidence of at least 0.9 that are wrong (CER$_{.9}$). The unpruned row is repeated in every column group. Cell colours mark the selector families.}
\label{tab:supp_grid_llava-v15-7b-b}
\end{table*}

\begin{table*}[t]
\centering
\footnotesize
\setlength{\tabcolsep}{3.4pt}
\renewcommand{\arraystretch}{0.88}
\begin{tabular}{@{}lcccccccc@{}}
\toprule
 & \multicolumn{4}{c}{keep 11\%} & \multicolumn{4}{c}{keep 22\%} \\
\cmidrule(lr){2-5}\cmidrule(lr){6-9}
Selector & Acc & ECE & AUROC & CER$_{.9}$ & Acc & ECE & AUROC & CER$_{.9}$ \\
\midrule
\multicolumn{9}{@{}l}{\cellcolor{hdrpope}\emph{POPE, first-token confidence}} \\
\cellcolor{hdrgrey}Unpruned & 87.1 & .032 & .830 & .047 & 87.1 & .032 & .830 & .047 \\
\cellcolor{famatt}CLS attention & 80.2 & .101 & .765 & .109 & 84.4 & .054 & .805 & .065 \\
\cellcolor{famatt}VisionZip & 80.2 & .099 & .783 & .104 & 84.7 & .048 & .809 & .061 \\
\cellcolor{famatt}FastV & 76.3 & .164 & .726 & .167 & 81.5 & .107 & .760 & .112 \\
\cellcolor{famcov}Coverage ($\alpha{=}0$) & 84.9 & .036 & .806 & .052 & 86.9 & .017 & .813 & .041 \\
\cellcolor{famcov}DivPrune & 86.3 & .019 & .808 & .040 & 87.8 & .017 & .806 & .041 \\
\cellcolor{famcov}CDPruner & 87.9 & .011 & .833 & .036 & 88.1 & .014 & .823 & .037 \\
\cellcolor{famrnd}Random & 78.9 & .087 & .788 & .086 & 83.1 & .052 & .796 & .067 \\
\midrule
\multicolumn{9}{@{}l}{\cellcolor{hdrgqa}\emph{GQA, probability of the whole answer}} \\
\cellcolor{hdrgrey}Unpruned & 63.3 & .060 & .809 & .069 & 63.3 & .060 & .809 & .069 \\
\cellcolor{famatt}CLS attention & 56.0 & .093 & .764 & .124 & 58.2 & .081 & .775 & .104 \\
\cellcolor{famatt}VisionZip & 56.1 & .095 & .768 & .122 & 57.8 & .085 & .773 & .109 \\
\cellcolor{famatt}FastV & 55.5 & .109 & .769 & .143 & 59.2 & .093 & .785 & .105 \\
\cellcolor{famcov}Coverage ($\alpha{=}0$) & 58.4 & .073 & .773 & .092 & 59.2 & .067 & .775 & .090 \\
\cellcolor{famcov}DivPrune & 57.3 & .072 & .778 & .095 & 58.8 & .064 & .776 & .084 \\
\cellcolor{famcov}CDPruner & 59.3 & .068 & .784 & .093 & 59.6 & .069 & .786 & .084 \\
\cellcolor{famrnd}Random & 55.6 & .092 & .764 & .128 & 57.8 & .080 & .770 & .112 \\
\midrule
\multicolumn{9}{@{}l}{\cellcolor{hdrgqa}\emph{GQA, first-token confidence}} \\
\cellcolor{hdrgrey}Unpruned & 63.3 & .072 & .801 & .079 & 63.3 & .072 & .801 & .079 \\
\cellcolor{famatt}CLS attention & 56.0 & .104 & .758 & .135 & 58.2 & .093 & .767 & .115 \\
\cellcolor{famatt}VisionZip & 56.1 & .105 & .763 & .131 & 57.8 & .096 & .765 & .121 \\
\cellcolor{famatt}FastV & 55.5 & .120 & .760 & .154 & 59.2 & .106 & .777 & .117 \\
\cellcolor{famcov}Coverage ($\alpha{=}0$) & 58.4 & .084 & .766 & .103 & 59.2 & .079 & .768 & .100 \\
\cellcolor{famcov}DivPrune & 57.3 & .083 & .772 & .105 & 58.8 & .076 & .769 & .097 \\
\cellcolor{famcov}CDPruner & 59.3 & .080 & .777 & .103 & 59.6 & .081 & .778 & .095 \\
\cellcolor{famrnd}Random & 55.6 & .102 & .758 & .135 & 57.8 & .091 & .764 & .121 \\
\bottomrule
\end{tabular}
\caption{LLaVA-1.5-13B: every selector and keep ratio. Accuracy in \%, ECE, AUROC, and the share of answers with confidence of at least 0.9 that are wrong (CER$_{.9}$). The unpruned row is repeated in every column group. Cell colours mark the selector families.}
\label{tab:supp_grid_llava-v15-13b}
\end{table*}

\begin{table*}[t]
\centering
\footnotesize
\setlength{\tabcolsep}{3.4pt}
\renewcommand{\arraystretch}{0.88}
\begin{tabular}{@{}lcccccccccccc@{}}
\toprule
 & \multicolumn{4}{c}{keep 11\%} & \multicolumn{4}{c}{keep 22\%} & \multicolumn{4}{c}{keep 33\%} \\
\cmidrule(lr){2-5}\cmidrule(lr){6-9}\cmidrule(lr){10-13}
Selector & Acc & ECE & AUROC & CER$_{.9}$ & Acc & ECE & AUROC & CER$_{.9}$ & Acc & ECE & AUROC & CER$_{.9}$ \\
\midrule
\multicolumn{13}{@{}l}{\cellcolor{hdrpope}\emph{POPE, first-token confidence}} \\
\cellcolor{hdrgrey}Unpruned & 87.9 & .067 & .826 & .067 & 87.9 & .067 & .826 & .067 & 87.9 & .067 & .826 & .067 \\
\cellcolor{famatt}CLS attention & 82.7 & .106 & .778 & .111 & 85.7 & .080 & .818 & .083 & 87.3 & .070 & .824 & .074 \\
\cellcolor{famatt}FastV & 79.5 & .149 & .748 & .157 & 85.0 & .094 & .801 & .095 & 87.2 & .077 & .807 & .082 \\
\cellcolor{famcov}Coverage ($\alpha{=}0$) & 83.7 & .086 & .821 & .081 & 85.9 & .074 & .822 & .076 & 87.2 & .068 & .823 & .069 \\
\cellcolor{famcov}DivPrune & 83.1 & .079 & .797 & .083 & 86.0 & .063 & .819 & .072 & 87.4 & .054 & .827 & .060 \\
\cellcolor{famrnd}Random & 81.0 & .084 & .763 & .098 & 84.1 & .078 & .792 & .087 & 85.2 & .078 & .810 & .083 \\
\midrule
\multicolumn{13}{@{}l}{\cellcolor{hdrgqa}\emph{GQA, probability of the whole answer}} \\
\cellcolor{hdrgrey}Unpruned & 60.1 & .134 & .798 & .140 & 60.1 & .134 & .798 & .140 & 60.1 & .134 & .798 & .140 \\
\cellcolor{famatt}CLS attention & 51.5 & .165 & .771 & .197 & 55.9 & .145 & .783 & .161 & 58.1 & .134 & .788 & .149 \\
\cellcolor{famatt}FastV & 48.5 & .202 & .748 & .252 & 53.3 & .181 & .771 & .201 & 55.7 & .163 & .786 & .176 \\
\cellcolor{famcov}Coverage ($\alpha{=}0$) & 55.7 & .124 & .780 & .148 & 57.8 & .124 & .791 & .142 & 58.6 & .123 & .791 & .136 \\
\cellcolor{famcov}DivPrune & 53.8 & .108 & .754 & .161 & 57.0 & .115 & .782 & .136 & 58.4 & .114 & .786 & .130 \\
\cellcolor{famrnd}Random & 52.3 & .059 & .726 & .177 & 55.4 & .085 & .749 & .169 & 57.3 & .109 & .766 & .152 \\
\midrule
\multicolumn{13}{@{}l}{\cellcolor{hdrgqa}\emph{GQA, first-token confidence}} \\
\cellcolor{hdrgrey}Unpruned & 60.1 & .151 & .785 & .158 & 60.1 & .151 & .785 & .158 & 60.1 & .151 & .785 & .158 \\
\cellcolor{famatt}CLS attention & 51.5 & .179 & .762 & .209 & 55.9 & .159 & .773 & .177 & 58.1 & .149 & .777 & .167 \\
\cellcolor{famatt}FastV & 48.5 & .216 & .739 & .270 & 53.3 & .196 & .761 & .218 & 55.7 & .178 & .775 & .192 \\
\cellcolor{famcov}Coverage ($\alpha{=}0$) & 55.7 & .138 & .769 & .164 & 57.8 & .139 & .778 & .158 & 58.6 & .139 & .779 & .152 \\
\cellcolor{famcov}DivPrune & 53.8 & .121 & .743 & .178 & 57.0 & .129 & .770 & .153 & 58.4 & .129 & .774 & .146 \\
\cellcolor{famrnd}Random & 52.3 & .060 & .717 & .193 & 55.4 & .101 & .736 & .184 & 57.3 & .124 & .754 & .172 \\
\midrule
\multicolumn{13}{@{}l}{\cellcolor{hdrgrey}\emph{ScienceQA-IMG, first-token confidence}} \\
\cellcolor{hdrgrey}Unpruned & 76.4 & .069 & .852 & .028 & 76.4 & .069 & .852 & .028 & 76.4 & .069 & .852 & .028 \\
\cellcolor{famatt}CLS attention & 75.2 & .039 & .866 & .012 & 74.5 & .055 & .866 & .019 & 74.4 & .067 & .865 & .019 \\
\cellcolor{famatt}FastV & 74.5 & .036 & .854 & .021 & 76.4 & .043 & .859 & .024 & 75.7 & .058 & .862 & .019 \\
\cellcolor{famcov}Coverage ($\alpha{=}0$) & 75.6 & .025 & .847 & .019 & 75.4 & .041 & .866 & .013 & 75.9 & .045 & .870 & .017 \\
\cellcolor{famcov}DivPrune & 75.2 & .029 & .853 & .013 & 74.6 & .047 & .873 & .018 & 75.6 & .043 & .859 & .019 \\
\cellcolor{famrnd}Random & 73.4 & .028 & .860 & .020 & 74.8 & .030 & .862 & .018 & 74.8 & .046 & .862 & .020 \\
\bottomrule
\end{tabular}
\caption{Qwen2.5-VL-3B: every selector and keep ratio. Accuracy in \%, ECE, AUROC, and the share of answers with confidence of at least 0.9 that are wrong (CER$_{.9}$). The unpruned row is repeated in every column group. Cell colours mark the selector families.}
\label{tab:supp_grid_qwen25-vl-3b}
\end{table*}

\begin{table*}[t]
\centering
\footnotesize
\setlength{\tabcolsep}{3.4pt}
\renewcommand{\arraystretch}{0.88}
\begin{tabular}{@{}lcccccccccccc@{}}
\toprule
 & \multicolumn{4}{c}{keep 11\%} & \multicolumn{4}{c}{keep 22\%} & \multicolumn{4}{c}{keep 33\%} \\
\cmidrule(lr){2-5}\cmidrule(lr){6-9}\cmidrule(lr){10-13}
Selector & Acc & ECE & AUROC & CER$_{.9}$ & Acc & ECE & AUROC & CER$_{.9}$ & Acc & ECE & AUROC & CER$_{.9}$ \\
\midrule
\multicolumn{13}{@{}l}{\cellcolor{hdrpope}\emph{POPE, first-token confidence}} \\
\cellcolor{hdrgrey}Unpruned & 87.8 & .082 & .822 & .083 & 87.8 & .082 & .822 & .083 & 87.8 & .082 & .822 & .083 \\
\cellcolor{famatt}CLS attention & 82.3 & .124 & .775 & .126 & 85.9 & .097 & .809 & .094 & 87.0 & .088 & .821 & .083 \\
\cellcolor{famatt}FastV & 77.3 & .185 & .728 & .189 & 83.5 & .126 & .771 & .125 & 85.6 & .101 & .806 & .100 \\
\cellcolor{famcov}Coverage ($\alpha{=}0$) & 84.3 & .097 & .802 & .092 & 86.3 & .084 & .809 & .090 & 87.2 & .082 & .804 & .084 \\
\cellcolor{famcov}DivPrune & 83.0 & .104 & .794 & .106 & 86.1 & .086 & .812 & .085 & 86.5 & .086 & .822 & .081 \\
\cellcolor{famrnd}Random & 80.4 & .132 & .768 & .138 & 83.9 & .107 & .793 & .109 & 85.8 & .094 & .799 & .095 \\
\midrule
\multicolumn{13}{@{}l}{\cellcolor{hdrgqa}\emph{GQA, probability of the whole answer}} \\
\cellcolor{hdrgrey}Unpruned & 61.0 & .200 & .803 & .190 & 61.0 & .200 & .803 & .190 & 61.0 & .200 & .803 & .190 \\
\cellcolor{famatt}CLS attention & 52.3 & .241 & .768 & .247 & 56.8 & .218 & .790 & .212 & 58.9 & .207 & .798 & .194 \\
\cellcolor{famatt}FastV & 47.2 & .286 & .746 & .311 & 53.5 & .254 & .779 & .250 & 56.2 & .235 & .786 & .229 \\
\cellcolor{famcov}Coverage ($\alpha{=}0$) & 57.7 & .197 & .777 & .202 & 59.3 & .199 & .794 & .194 & 60.4 & .195 & .793 & .191 \\
\cellcolor{famcov}DivPrune & 56.1 & .187 & .775 & .195 & 58.8 & .193 & .788 & .192 & 59.5 & .198 & .797 & .188 \\
\cellcolor{famrnd}Random & 54.2 & .180 & .750 & .227 & 57.3 & .194 & .770 & .208 & 58.9 & .199 & .784 & .200 \\
\midrule
\multicolumn{13}{@{}l}{\cellcolor{hdrgqa}\emph{GQA, first-token confidence}} \\
\cellcolor{hdrgrey}Unpruned & 61.0 & .216 & .788 & .206 & 61.0 & .216 & .788 & .206 & 61.0 & .216 & .788 & .206 \\
\cellcolor{famatt}CLS attention & 52.3 & .254 & .759 & .262 & 56.8 & .232 & .778 & .228 & 58.9 & .221 & .784 & .211 \\
\cellcolor{famatt}FastV & 47.2 & .299 & .734 & .327 & 53.5 & .267 & .768 & .267 & 56.2 & .249 & .775 & .244 \\
\cellcolor{famcov}Coverage ($\alpha{=}0$) & 57.7 & .212 & .764 & .220 & 59.3 & .213 & .780 & .212 & 60.4 & .209 & .779 & .208 \\
\cellcolor{famcov}DivPrune & 56.1 & .202 & .765 & .208 & 58.8 & .208 & .775 & .207 & 59.5 & .212 & .785 & .204 \\
\cellcolor{famrnd}Random & 54.2 & .197 & .737 & .243 & 57.3 & .210 & .757 & .224 & 58.9 & .214 & .771 & .218 \\
\midrule
\multicolumn{13}{@{}l}{\cellcolor{hdrgrey}\emph{ScienceQA-IMG, first-token confidence}} \\
\cellcolor{hdrgrey}Unpruned & 84.3 & .059 & .896 & .041 & 84.3 & .059 & .896 & .041 & 84.3 & .059 & .896 & .041 \\
\cellcolor{famatt}CLS attention & 79.2 & .066 & .869 & .041 & 79.9 & .077 & .882 & .044 & 83.7 & .052 & .880 & .033 \\
\cellcolor{famatt}FastV & 75.7 & .077 & .840 & .041 & 80.9 & .066 & .879 & .042 & 82.1 & .070 & .885 & .049 \\
\cellcolor{famcov}Coverage ($\alpha{=}0$) & 79.2 & .057 & .878 & .041 & 81.5 & .055 & .885 & .028 & 81.8 & .060 & .895 & .028 \\
\cellcolor{famcov}DivPrune & 76.8 & .079 & .875 & .041 & 80.6 & .059 & .880 & .031 & 81.6 & .058 & .893 & .028 \\
\cellcolor{famrnd}Random & 76.6 & .077 & .864 & .053 & 80.1 & .057 & .866 & .039 & 80.5 & .065 & .887 & .034 \\
\bottomrule
\end{tabular}
\caption{Qwen2.5-VL-7B: every selector and keep ratio. Accuracy in \%, ECE, AUROC, and the share of answers with confidence of at least 0.9 that are wrong (CER$_{.9}$). The unpruned row is repeated in every column group. Cell colours mark the selector families.}
\label{tab:supp_grid_qwen25-vl-7b}
\end{table*}

\begin{table*}[t]
\centering
\footnotesize
\setlength{\tabcolsep}{3.4pt}
\renewcommand{\arraystretch}{0.88}
\begin{tabular}{@{}lcccccccc@{}}
\toprule
 & \multicolumn{4}{c}{keep 11\%} & \multicolumn{4}{c}{keep 22\%} \\
\cmidrule(lr){2-5}\cmidrule(lr){6-9}
Selector & Acc & ECE & AUROC & CER$_{.9}$ & Acc & ECE & AUROC & CER$_{.9}$ \\
\midrule
\multicolumn{9}{@{}l}{\cellcolor{hdrpope}\emph{POPE, first-token confidence}} \\
\cellcolor{hdrgrey}Unpruned & 86.8 & .085 & .810 & .083 & 86.8 & .085 & .810 & .083 \\
\cellcolor{famatt}CLS attention & 78.7 & .151 & .748 & .162 & 84.0 & .103 & .799 & .103 \\
\cellcolor{famatt}FastV & 64.8 & .265 & .721 & .266 & 75.7 & .188 & .760 & .184 \\
\cellcolor{famcov}Coverage ($\alpha{=}0$) & 81.8 & .134 & .787 & .129 & 83.5 & .109 & .809 & .103 \\
\cellcolor{famcov}DivPrune & 80.6 & .136 & .775 & .134 & 84.4 & .102 & .792 & .105 \\
\cellcolor{famrnd}Random & 78.2 & .166 & .771 & .163 & 82.3 & .125 & .787 & .129 \\
\midrule
\multicolumn{9}{@{}l}{\cellcolor{hdrgqa}\emph{GQA, probability of the whole answer}} \\
\cellcolor{hdrgrey}Unpruned & 59.9 & .161 & .801 & .165 & 59.9 & .161 & .801 & .165 \\
\cellcolor{famatt}CLS attention & 51.2 & .206 & .771 & .226 & 55.5 & .181 & .787 & .194 \\
\cellcolor{famatt}FastV & 44.1 & .204 & .735 & .297 & 49.5 & .217 & .760 & .249 \\
\cellcolor{famcov}Coverage ($\alpha{=}0$) & 55.5 & .182 & .788 & .191 & 57.5 & .171 & .798 & .168 \\
\cellcolor{famcov}DivPrune & 53.1 & .194 & .781 & .201 & 56.8 & .174 & .795 & .177 \\
\cellcolor{famrnd}Random & 52.0 & .203 & .775 & .226 & 55.4 & .188 & .791 & .192 \\
\midrule
\multicolumn{9}{@{}l}{\cellcolor{hdrgqa}\emph{GQA, first-token confidence}} \\
\cellcolor{hdrgrey}Unpruned & 59.9 & .186 & .785 & .191 & 59.9 & .186 & .785 & .191 \\
\cellcolor{famatt}CLS attention & 51.2 & .226 & .760 & .242 & 55.5 & .204 & .773 & .214 \\
\cellcolor{famatt}FastV & 44.1 & .222 & .726 & .312 & 49.5 & .239 & .749 & .267 \\
\cellcolor{famcov}Coverage ($\alpha{=}0$) & 55.5 & .208 & .770 & .219 & 57.5 & .196 & .782 & .192 \\
\cellcolor{famcov}DivPrune & 53.1 & .218 & .765 & .228 & 56.8 & .197 & .780 & .201 \\
\cellcolor{famrnd}Random & 52.0 & .230 & .759 & .250 & 55.4 & .215 & .773 & .221 \\
\bottomrule
\end{tabular}
\caption{Qwen2.5-VL-32B: every selector and keep ratio. Accuracy in \%, ECE, AUROC, and the share of answers with confidence of at least 0.9 that are wrong (CER$_{.9}$). The unpruned row is repeated in every column group. Cell colours mark the selector families.}
\label{tab:supp_grid_qwen25-vl-32b}
\end{table*}

\section{Risk Control}
\label{supp:risk}

\paragraph{Monotone envelope.}
Let $L_j(\tau)\in[0,1]$ be the average deviation over the questions of image $j$ at threshold $\tau$.
The threshold ranges over the finite grid $\Lambda=\{\sigma(z): z=-5,-4.98,\dots,15\}\cup\{\tau_{\max}\}$, with $\sigma$ the logistic function, fixed before seeing any data.
The spacing in $z$ resolves confidences close to one, which are common for Qwen2.5-VL.
The value $\tau_{\max}$ exceeds every confidence, so that every question reaches the unpruned model and $L_j(\tau_{\max})=0$.
On a finite grid the minimum in Eq.~(3) of the main paper is attained, and a question exits early at stage $m$ exactly when $c_m\ge\hat\tau$.
The envelope $\tilde L_j(\tau)=\max_{\tau'\in\Lambda,\,\tau'\ge\tau}L_j(\tau')$ is non-increasing on $\Lambda$, satisfies $\tilde L_j\ge L_j$, and is zero at $\tau_{\max}$.
Corollary~1 of \citet{angelopoulos2024crc}, which introduces this envelope for non-monotone losses, then gives $\mathbb{E}[L_{n+1}(\hat\tau)]\le\varepsilon$ for exchangeable images with loss bound $B{=}1$.
The guaranteed quantity is the expected deviation of a new image averaged over its questions.
On POPE every image has 18 questions and on ScienceQA-IMG one, so this equals the deviation averaged over questions; on GQA the two differ, and we report the image average.

\end{document}